\documentclass[fleqn,twoside]{article}
\usepackage{espcrc1}

\newcommand{\AmS}{{\protect\the\textfont2
  A\kern-.1667em\lower.5ex\hbox{M}\kern-.125emS}}

\title{Towards Detection of Bottlenecks in Modular Systems}

\author{Mark Sh. Levin
\thanks{
 Mark Sh. Levin:~
  http://www.mslevin.iitp.ru;
 email: mslevin@acm.org
  }
  }

\begin{document}

\maketitle

\begin{abstract}
 The paper describes some basic approaches to
 detection of bottlenecks in composite (modular) systems.
%
%
 The following basic system bottlenecks detection problems are examined:
 (1) traditional quality management approaches
  (Pareto chart based method,
  multicriteria analysis
  as selection of Pareto-efficient points,
 and/or multicriteria ranking),
 (2) selection of critical system elements
%
  (critical components/modules,
 critical component interconnection),
%
%
 (3) selection of interconnected system components
 as composite system faults
 (via clique-based fusion),
 (4) critical elements (e.g., nodes) in networks,
 and
%
%
 (5) predictive detection of system bottlenecks
 (detection of system components based on forecasting of their
 parameters).
 Here, heuristic solving schemes are used.
 Numerical examples illustrate the approaches.

~~


\vspace{1pc}
\end{abstract}



\newcounter{cms}
\setlength{\unitlength}{1mm}

\section{Introduction}

  In recent decades,
 the significance of modular (multi-component) systems has been increased
 (e.g., \cite{bald00,dah01,garud09,hua98,jose05,lev98,lev06,lev12morph,lev13intro,ulrich94,voss09}).
 This paper describes approaches to
 detection of bottlenecks in composite (modular) systems.
 Here, the following is assumed (Fig. 1):

 {\it 1.} The considered hierarchical modular system can be represented as
 a morphological structure
 (e.g., \cite{lev98,lev06,lev12morph,lev12hier})
 or as a network.

 {\it 2.} The following system elements are under examination as
 the bottlenecks:
 (i) system component (or a system component fault),
 (ii) interconnection between system components (compatibility),
 (iii) group of system components (or a group of system faults),
 (iv) group of interconnected system components
 (or a composite system faults).
 Thus, the system bottlenecks are considered as
 low quality system part(s)/component(s)
 or system fault(s) and their compositions.

 In the paper, the following approaches are described
 (Table 1):

 {\bf I.} Basic quality management approaches:~
  (1.1.) Pareto chart based method \cite{ishikawa85},
%
  (1.2.) multicriteria analysis
  as selection of Pareto-efficient points,
 and/or multicriteria ranking \cite{lev88a,lev06}).

 {\bf II.} Detection of low quality system parts:~
 (2.1.) detection of critical components/modules,
 (2.2.) detection of critical component interconnection
 (component compatibility),
 and
 (2.3.) analysis of the system structure and
 detecting the situation when
 the system structure has to be improved.

 {\bf III.} Selection of interconnected system components
 as composite system faults
 (e.g., via hierarchical morphological design
 \cite{lev98,lev06,lev12morph,lev12a},
  via clique-based fusion \cite{lev12clique,levlast06}).

 {\bf IV.} Detection of critical components in networks,
 detection of low quality node interconnection,
  detection of low quality network topology,
 e.g., definition of the internal network nodes via
 maximum leaf spanning tree problem
 (e.g., \cite{alon09,caro00,gar79}),
  connected dominating sets problem
  (e.g., \cite{blum05,caro00,gar79}),
 hierarchical  network design problem
 (e.g., \cite{bala94,current86,pirkul91}).

 {\bf V.} Predictive detection of system bottlenecks:
  (5.1.) predictive detection of system components based on forecasting of their
 parameters,
 (5.2.) predictive detection of
 of critical components in networks,
  low quality node interconnection,
  low quality network topology,
 (5.3.) predictive detection of group
 of interconnected system components based on
 clique based fusion of graph streams
 \cite{lev12clique}.

  Mainly, composite engineering frameworks (i.e., heuristics solving schemes) are used.
 Numerical examples illustrate the approaches.

\begin{center}
\begin{picture}(90,44)
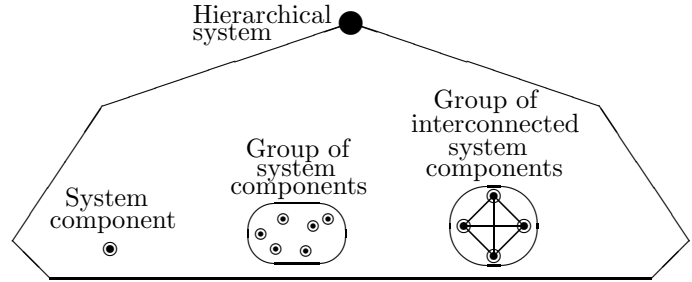
\put(03,00){\makebox(0,0)[bl] {Fig. 1. Illustration for systems
 and system bottlenecks}}

\put(45,40){\circle*{3}}

\put(24,40){\makebox(0,0)[bl]{Hierarchical}}
\put(24,37){\makebox(0,0)[bl]{system}}


\put(07,15){\makebox(0,0)[bl]{System}}
\put(05,12){\makebox(0,0)[bl]{component}}

\put(13,10){\circle*{1.2}} \put(13,10){\circle{1.9}}


\put(56,28){\makebox(0,0)[bl]{Group of}}
\put(53,25.5){\makebox(0,0)[bl]{interconnected}}
\put(58,21.8){\makebox(0,0)[bl]{system}}
\put(55,19){\makebox(0,0)[bl]{components}}

\put(68,13){\circle*{1.2}} \put(68,13){\circle{1.9}}
\put(60,13){\circle*{1.2}} \put(60,13){\circle{1.9}}
\put(64,17){\circle*{1.2}} \put(64,17){\circle{1.9}}
\put(64,09){\circle*{1.2}} \put(64,09){\circle{1.9}}

\put(60,13){\line(1,0){08}} \put(64,09){\line(0,1){08}}
\put(60,13){\line(1,1){04}} \put(68,13){\line(-1,1){04}}
\put(64,09){\line(-1,1){04}} \put(64,09){\line(1,1){04}}

\put(64,13){\oval(11.6,10.5)}


\put(31,21.5){\makebox(0,0)[bl]{Group of}}
\put(32.5,19){\makebox(0,0)[bl]{system}}
\put(29,16.5){\makebox(0,0)[bl]{components}}

\put(39,09.6){\circle*{0.8}} \put(39,09.6){\circle{1.4}}
\put(42,14){\circle*{0.8}} \put(42,14){\circle{1.4}}
\put(35,10){\circle*{0.8}} \put(35,10){\circle{1.4}}
\put(36,14){\circle*{0.8}} \put(36,14){\circle{1.4}}
\put(33,12){\circle*{0.8}} \put(33,12){\circle{1.4}}
\put(40,13){\circle*{0.8}} \put(40,13){\circle{1.4}}

\put(37.8,12){\oval(13,08)}


\put(05,06){\line(1,0){80}}

\put(05,06){\line(-1,1){5}} \put(85,06){\line(1,1){5}}

\put(00,11){\line(2,3){12}} \put(12,29){\line(3,1){33}}

\put(90,11){\line(-2,3){12}} \put(78,29){\line(-3,1){33}}


\end{picture}
\end{center}

\begin{center}
\begin{picture}(116,143)
\put(10,139){\makebox(0,0)[bl] {Table 1. Basic approaches to
detection of system bottlenecks}}

\put(00,00){\line(1,0){116}}  \put(00,127){\line(1,0){116}}
\put(00,137){\line(1,0){116}}

\put(00,00){\line(0,1){137}}  \put(30,00){\line(0,1){137}}
\put(62,00){\line(0,1){137}}  \put(116,00){\line(0,1){137}}

\put(01,132.5){\makebox(0,0)[bl] {Objects under }}
\put(01,129.4){\makebox(0,0)[bl] {examination}}

\put(31,133){\makebox(0,0)[bl] {Basic detection }}
\put(31,129){\makebox(0,0)[bl] {methods/models}}

\put(63,133){\makebox(0,0)[bl] {Predictive detection}}
\put(63,129){\makebox(0,0)[bl] {(forecasting/dynamics)}}


\put(00,122){\makebox(0,0)[bl]{1.System compo-}}
\put(03,118){\makebox(0,0)[bl]{nent (or system}}
\put(03,114){\makebox(0,0)[bl]{component fault)}}


\put(31,122){\makebox(0,0)[bl]{(a) Pareto chart}}
\put(31,118){\makebox(0,0)[bl]{method \cite{ishikawa85}}}

\put(31,114){\makebox(0,0)[bl]{(b) multicriteria}}
\put(31,110){\makebox(0,0)[bl]{analysis/ranking }}
\put(31,106){\makebox(0,0)[bl]{(sorting) \cite{lev88a,lev06}}}


\put(63,122){\makebox(0,0)[bl]{(a) Pareto chart method based }}
\put(63,118){\makebox(0,0)[bl]{on forecast of system components}}
\put(63,114){\makebox(0,0)[bl]{parameters}}

\put(63,110){\makebox(0,0)[bl]{(b) multicriteria
 analysis/ranking}}

\put(63,106){\makebox(0,0)[bl]{(sorting) based on forecast of}}
\put(63,102){\makebox(0,0)[bl]{system component parameters}}

\put(01,97){\makebox(0,0)[bl]{2.Interconnection}}
\put(03,93.5){\makebox(0,0)[bl]{of system}}
\put(03,89.5){\makebox(0,0)[bl]{components}}


\put(31,96.5){\makebox(0,0)[bl]{(a) Pareto chart}}
\put(31,93){\makebox(0,0)[bl]{method \cite{ishikawa85}}}

\put(31,89){\makebox(0,0)[bl]{(b) multicriteria}}
\put(31,85){\makebox(0,0)[bl]{analysis/ranking}}
\put(31,81){\makebox(0,0)[bl]{(sorting)}}


\put(63,97){\makebox(0,0)[bl]{(a) Pareto- chart method based }}
\put(63,93){\makebox(0,0)[bl]{on forecast of system components}}
\put(63,89){\makebox(0,0)[bl]{parameters}}

\put(63,85){\makebox(0,0)[bl]{(b) multicriteria analysis/ranking}}
\put(63,81){\makebox(0,0)[bl]{(sorting) based on forecast of}}
\put(63,77){\makebox(0,0)[bl]{system component parameters}}


\put(01,72){\makebox(0,0)[bl]{3.Group of system}}
\put(03,68){\makebox(0,0)[bl]{components (or}}
\put(03,64){\makebox(0,0)[bl]{composite fault)}}

\put(31,72){\makebox(0,0)[bl]{Multicriteria}}
\put(31,68){\makebox(0,0)[bl]{analysis (sorting)}}
\put(31,64){\makebox(0,0)[bl]{\cite{lev88a,lev12b,roy96,zap02}}}

\put(63,72){\makebox(0,0)[bl]{Multicriteria analysis (sorting)}}
\put(63,68.7){\makebox(0,0)[bl]{based on forecast of network}}
\put(63,64){\makebox(0,0)[bl]{parameters
 \cite{lev88a,lev12b,roy96,zap02}}}


\put(00.7,59){\makebox(0,0)[bl]{4.Bottlenecks (e.g.,}}
\put(03,55){\makebox(0,0)[bl]{critical nodes)}}
\put(03,51.7){\makebox(0,0)[bl]{in networks}}


\put(31,59){\makebox(0,0)[bl]{(a) maximum  leaf}}
\put(31,55){\makebox(0,0)[bl]{spanning tree }}
\put(31,51){\makebox(0,0)[bl]{\cite{alon09,caro00,gar79}}}

\put(31,47){\makebox(0,0)[bl]{(b) connected }}
\put(31,43){\makebox(0,0)[bl]{dominating set }}

\put(31,39){\makebox(0,0)[bl]{\cite{blum05,caro00,gar79}}}

\put(31,35){\makebox(0,0)[bl]{(c) hierarchical}}
\put(31,31){\makebox(0,0)[bl]{network design}}
\put(31,27){\makebox(0,0)[bl]{\cite{bala94,current86,pirkul91}}}

\put(31,22.5){\makebox(0,0)[bl]{(d) low-quality no-}}
\put(31,19.5){\makebox(0,0)[bl]{de interconnection}}


\put(63,59){\makebox(0,0)[bl]{(a) maximum leaf spanning tree}}
\put(63,55.7){\makebox(0,0)[bl]{based on forecast of network}}

\put(63,51){\makebox(0,0)[bl]{(b) connected dominating set}}
\put(63,47.7){\makebox(0,0)[bl]{based on forecast of network}}

\put(63,43){\makebox(0,0)[bl]{(c) hierarchical network design}}
\put(63,39.7){\makebox(0,0)[bl]{based on forecast of network}}

\put(63,35){\makebox(0,0)[bl]{(d) low-quality node interconnec-}}
\put(63,31.7){\makebox(0,0)[bl]{tion based on forecast of
network}}


\put(01,14){\makebox(0,0)[bl]{5.Group of inter-}}
\put(03,10){\makebox(0,0)[bl]{connected sys-}}
\put(03,06){\makebox(0,0)[bl]{tem components}}

\put(31,14){\makebox(0,0)[bl]{(a) HMMD}}
\put(31,10){\makebox(0,0)[bl]{\cite{lev98,lev06,lev12morph,lev12a}}}

\put(31,06){\makebox(0,0)[bl]{(b) clique-based}}
\put(31,02){\makebox(0,0)[bl]{fusion
 \cite{lev12clique,levlast06}}}

\put(63,14){\makebox(0,0)[bl]{(a) HMMD based on system}}
\put(63,10){\makebox(0,0)[bl]{forecast
\cite{lev98,lev06,lev12morph,lev12a}}}

\put(63,06){\makebox(0,0)[bl]{(b) clique-based fusion over}}
\put(63,02){\makebox(0,0)[bl]{graph streams
 \cite{lev12clique}}}



\end{picture}
\end{center}

\section{Traditional Quality Control Methods}


 Here, two methods for the analysis of system components are considered:
 (i) Pareto chart based method,
 (ii) multicriteria analysis (ranking).
 The first method is the main method to the detection of system bottlenecks
 in Japanese approach of quality control and consists in
 the analysis of system components/parts by their reliability
 (or the frequency of component fault/failure/trouble/anomaly)
 (e.g., \cite{ishikawa85}):

 {\it Step 1.} Definition of the initial set of system
 components/part for the analysis.

 {\it Step 2.} Assessment of reliability
 (i.e., frequency of the component fault.

 {\it Step 3.} Selection of the non-reliable
 system components (as system bottlenecks)
 by a threshold (at the Pareto chart).


 Further, the method  based on multicriteria description and
 multicriteria analysis of the system components
 has been suggested in
 \cite{lev88a,lev06}:

  {\it Step 1.} Definition of the initial set of system
 components/part for the analysis.

 {\it Step 2.} Assessment of the system
 components by many criteria.

 {\it Step 3.} Multicriteria ranking
 (e.g., selection of Pareto-efficient elements, outranking
 techniques, utility function analysis)
 of the system components to select the most important ones
 from the viewpoint of the total system safety
 (as system bottlenecks).

 Now, let us consider an illustrative example:
 supercharger for gas-pump aggregate
%
  \cite{lev88a}.
 The tree-like structure of the considered aggregate
 is the following (Fig. 2):

  {\bf 1.} Body frame components:~
%
 {\it 1.1.} external body,
 {\it 1.2.}
 body cover,
%
 {\it 1.3.}
 internal body with
 embedded elements,
 and
%
 {\it 1.4.}
 body seal.

 {\bf 2.} Supporting block bearers.

 {\bf 3.} Oil seals.

 {\bf 4.} Rotor.

 {\bf 5.} Connection units:
%
 {\it 5.1.} half-clutch,
%
 {\it 5.2.} gear hoop,
 and
%
 {\it 5.3.} torsion shaft.

 {\bf 6.}  system of lubrication:
%
 {\it 6.1.} oil boiler,
%
 {\it 6.2.} oil filters,
%
 {\it 6.3.} main oil pump,
%
 {\it 6.4.} start oil pump,
%
 {\it 6.5.} armature,
%
 {\it 6.6.} valve elements,
%
 {\it 6.7.} temperature regulator,
%
 {\it 6.8.} oil coolers,
 and
%
 {\it 6.9.} for oil coolers.

 {\bf 7.} System of oil seals:~
%
 {\it 7.1.} oil boiler,
%
 {\it 7.2.} oil filter,
%
 {\it 7.3.} main pump,
%
 {\it 7.4.} start pump,
%
 {\it 7.5} pressure regulator,
%
 {\it 7.6.} hydro-accumulator,
%
 {\it 7.7.} stripping vessel,
%
 {\it 7.8.} oil deriving,
%
 {\it 7.9.} pipelines,
%
 {\it 7.10.} valve elements,
%
 and
 {\it 7.11.} gum elastic seal rings.

 {\bf 8.} Thrust blocks:
%
 {\it 8.1.} pad,
%
 {\it 8.2.} wrapper rings,
 {\it 8.3.} stop rings,
 and
%
 {\it 8.4.} distance rings.

\begin{center}
\begin{picture}(110,90)
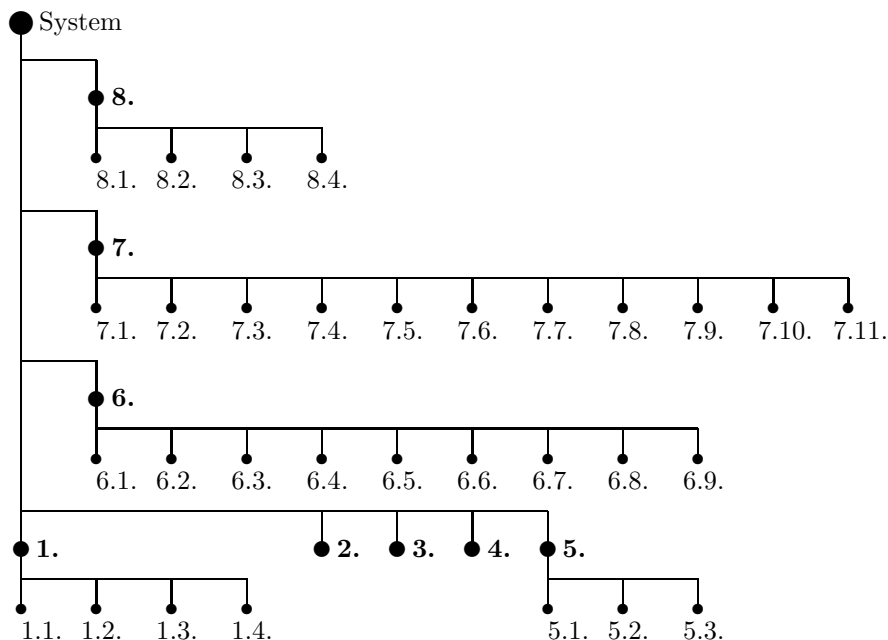
\put(23,00){\makebox(0,0)[bl] {Fig. 2. Structure of the examined
 system}}

\put(00,87){\circle*{3}}

\put(2.4,85.4){\makebox(0,0)[bl]{System}}


\put(00,82){\line(1,0){10}}

\put(10,77){\circle*{2}} \put(10,73){\line(0,1){9}}

\put(12,76){\makebox(0,0)[bl]{{\bf 8.}}}

\put(10,73){\line(1,0){30}}

\put(10,69){\circle*{1.5}} \put(10,69){\line(0,1){04}}
\put(20,69){\circle*{1.5}} \put(20,69){\line(0,1){04}}
\put(30,69){\circle*{1.5}} \put(30,69){\line(0,1){04}}
\put(40,69){\circle*{1.5}} \put(40,69){\line(0,1){04}}

\put(10,65){\makebox(0,0)[bl]{\(8.1.\)}}
\put(18,65){\makebox(0,0)[bl]{\(8.2.\)}}
\put(28,65){\makebox(0,0)[bl]{\(8.3.\)}}
\put(38,65){\makebox(0,0)[bl]{\(8.4.\)}}


\put(00,62){\line(1,0){10}}

\put(10,57){\circle*{2}} \put(10,53){\line(0,1){9}}

\put(12,56){\makebox(0,0)[bl]{{\bf 7.}}}

\put(10,53){\line(1,0){100}}

\put(10,49){\circle*{1.5}} \put(10,49){\line(0,1){04}}
\put(20,49){\circle*{1.5}} \put(20,49){\line(0,1){04}}
\put(30,49){\circle*{1.5}} \put(30,49){\line(0,1){04}}
\put(40,49){\circle*{1.5}} \put(40,49){\line(0,1){04}}
\put(50,49){\circle*{1.5}} \put(50,49){\line(0,1){04}}
\put(60,49){\circle*{1.5}} \put(60,49){\line(0,1){04}}
\put(70,49){\circle*{1.5}} \put(70,49){\line(0,1){04}}
\put(80,49){\circle*{1.5}} \put(80,49){\line(0,1){04}}
\put(90,49){\circle*{1.5}} \put(90,49){\line(0,1){04}}
\put(100,49){\circle*{1.5}} \put(100,49){\line(0,1){04}}
\put(110,49){\circle*{1.5}} \put(110,49){\line(0,1){04}}

\put(10,45){\makebox(0,0)[bl]{\(7.1.\)}}
\put(18,45){\makebox(0,0)[bl]{\(7.2.\)}}
\put(28,45){\makebox(0,0)[bl]{\(7.3.\)}}
\put(38,45){\makebox(0,0)[bl]{\(7.4.\)}}
\put(48,45){\makebox(0,0)[bl]{\(7.5.\)}}
\put(58,45){\makebox(0,0)[bl]{\(7.6.\)}}
\put(68,45){\makebox(0,0)[bl]{\(7.7.\)}}
\put(78,45){\makebox(0,0)[bl]{\(7.8.\)}}
\put(88,45){\makebox(0,0)[bl]{\(7.9.\)}}
\put(98,45){\makebox(0,0)[bl]{\(7.10.\)}}
\put(108,45){\makebox(0,0)[bl]{\(7.11.\)}}


\put(00,42){\line(1,0){10}}

\put(10,37){\circle*{2}} \put(10,33){\line(0,1){9}}

\put(12,36){\makebox(0,0)[bl]{{\bf 6.}}}

\put(10,33){\line(1,0){80}}

\put(10,29){\circle*{1.5}} \put(10,29){\line(0,1){04}}
\put(20,29){\circle*{1.5}} \put(20,29){\line(0,1){04}}
\put(30,29){\circle*{1.5}} \put(30,29){\line(0,1){04}}
\put(40,29){\circle*{1.5}} \put(40,29){\line(0,1){04}}
\put(50,29){\circle*{1.5}} \put(50,29){\line(0,1){04}}
\put(60,29){\circle*{1.5}} \put(60,29){\line(0,1){04}}
\put(70,29){\circle*{1.5}} \put(70,29){\line(0,1){04}}
\put(80,29){\circle*{1.5}} \put(80,29){\line(0,1){04}}
\put(90,29){\circle*{1.5}} \put(90,29){\line(0,1){04}}

\put(10,25){\makebox(0,0)[bl]{\(6.1.\)}}
\put(18,25){\makebox(0,0)[bl]{\(6.2.\)}}
\put(28,25){\makebox(0,0)[bl]{\(6.3.\)}}
\put(38,25){\makebox(0,0)[bl]{\(6.4.\)}}
\put(48,25){\makebox(0,0)[bl]{\(6.5.\)}}
\put(58,25){\makebox(0,0)[bl]{\(6.6.\)}}
\put(68,25){\makebox(0,0)[bl]{\(6.7.\)}}
\put(78,25){\makebox(0,0)[bl]{\(6.8.\)}}
\put(88,25){\makebox(0,0)[bl]{\(6.9.\)}}


\put(70,17){\circle*{2}} \put(70,13){\line(0,1){9}}

\put(72,16){\makebox(0,0)[bl]{{\bf 5.}}}

\put(70,13){\line(1,0){20}}

\put(70,09){\circle*{1.5}} \put(70,09){\line(0,1){04}}
\put(80,09){\circle*{1.5}} \put(80,09){\line(0,1){04}}
\put(90,09){\circle*{1.5}} \put(90,09){\line(0,1){04}}

\put(70,05){\makebox(0,0)[bl]{\(5.1.\)}}
\put(78,05){\makebox(0,0)[bl]{\(5.2.\)}}
\put(88,05){\makebox(0,0)[bl]{\(5.3.\)}}


\put(00,22){\line(1,0){70}}


\put(60,17){\line(0,1){5}} \put(60,17){\circle*{2}}

\put(62,16){\makebox(0,0)[bl]{{\bf 4.}}}


\put(50,17){\line(0,1){5}} \put(50,17){\circle*{2}}

\put(52,16){\makebox(0,0)[bl]{{\bf 3.}}}


\put(40,17){\line(0,1){5}} \put(40,17){\circle*{2}}

\put(42,16){\makebox(0,0)[bl]{{\bf 2.}}}


\put(00,13){\line(0,1){74}} \put(00,17){\circle*{2}}

\put(02,16){\makebox(0,0)[bl]{{\bf 1.}}}

\put(00,13){\line(1,0){30}}

\put(00,09){\circle*{1.5}} \put(00,09){\line(0,1){04}}
\put(10,09){\circle*{1.5}} \put(10,09){\line(0,1){04}}
\put(20,09){\circle*{1.5}} \put(20,09){\line(0,1){04}}
\put(30,09){\circle*{1.5}} \put(30,09){\line(0,1){04}}

\put(00,05){\makebox(0,0)[bl]{\(1.1.\)}}
\put(08,05){\makebox(0,0)[bl]{\(1.2.\)}}
\put(18,05){\makebox(0,0)[bl]{\(1.3.\)}}
\put(28,05){\makebox(0,0)[bl]{\(1.4.\)}}

\end{picture}
\end{center}

 The following six criteria
 (local, systemic) are examined:~
 {\it 1.}
 \(C_{1}\), frequency of faults (percent);
 {\it 2.}
 \(C_{2}\),
 time of
 ``out of work''
 in the case of the component fault;
 {\it 3.}
 \(C_{3}\),
 cost of work to repair
  the apparatus;
 {\it 4.}
 \(C_{4}\),
 level of influence of component fault to
 other system components, scale \([0,1,2]\)
 (no influence: \(0\),
 influence exists: \(1\),
 strong influence: \(2\));
 {\it 5.}
 \(C_{5}\),
 wideness of usage, scale \([0,1,2]\)
 (the component is used
 in the only this apparatus: \(0\),
 the component is used in other apparatus: \(1\),
 the component is used in many various systems: \(2\));
 {\it 6.}
 \(C_{6}\), level of influence of component fault to
 total system safety, scale \([0,1]\)
 (no influence: \(0\),
 the influence exists: \(1\)).

 Table 2 contains multicriteria description
 (i.e., estimates upon the considered six criteria)
 of the considered pump system
  (statistical data, processing, and expert judgment)
 \cite{lev88a}.

 The selection of system bottlenecks by Pareto chart is
 illustrated in Fig. 3 (estimates upon criterion \(C_{1}\)):
 (a) threshold 1 (6.8), system bottlenecks components are the
 following:~
  4, 7.11;
  (b) threshold 2 (1.5), system bottlenecks components are the
 following:~
 2, 4, 6.3, 6.8, 7.5, 7.7, 7.11.

\begin{center}
\begin{picture}(82,160)

\put(08,156){\makebox(0,0)[bl]{Table 2. Estimates of system
 components}}

\put(00,0){\line(1,0){82}} \put(00,142){\line(1,0){82}}
\put(22,148){\line(1,0){60}} \put(00,154){\line(1,0){82}}

\put(00,00){\line(0,1){154}} \put(22,00){\line(0,1){154}}

\put(32,00){\line(0,1){148}} \put(42,00){\line(0,1){148}}
\put(52,00){\line(0,1){148}} \put(62,00){\line(0,1){148}}
\put(72,00){\line(0,1){148}}

\put(82,00){\line(0,1){154}}


\put(01,149.5){\makebox(0,0)[bl]{System part/}}
\put(01,145.5){\makebox(0,0)[bl]{component}}

\put(34,149.5){\makebox(0,0)[bl]{Estimates upon criteria}}

\put(24,144){\makebox(0,0)[bl]{\(C_{1}\)}}
\put(34,144){\makebox(0,0)[bl]{\(C_{2}\)}}
\put(44,144){\makebox(0,0)[bl]{\(C_{3}\)}}
\put(54,144){\makebox(0,0)[bl]{\(C_{4}\)}}
\put(64,144){\makebox(0,0)[bl]{\(C_{5}\)}}
\put(74,144){\makebox(0,0)[bl]{\(C_{6}\)}}



\put(01,138){\makebox(0,0)[bl]{1.1.}}

\put(24,138){\makebox(0,0)[bl]{\(0.25\)}}
\put(34,138){\makebox(0,0)[bl]{\(12.8\)}}
\put(44,138){\makebox(0,0)[bl]{\(18.0\)}}

\put(56,138){\makebox(0,0)[bl]{\(0\)}}
\put(66,138){\makebox(0,0)[bl]{\(0\)}}
\put(76,138){\makebox(0,0)[bl]{\(1\)}}


\put(01,134){\makebox(0,0)[bl]{1.2.}}

\put(24,134){\makebox(0,0)[bl]{\(0.00\)}}
\put(34,134){\makebox(0,0)[bl]{\(12.8\)}}
\put(44,134){\makebox(0,0)[bl]{\(18.0\)}}

\put(56,134){\makebox(0,0)[bl]{\(0\)}}
\put(66,134){\makebox(0,0)[bl]{\(0\)}}
\put(76,134){\makebox(0,0)[bl]{\(1\)}}


\put(01,130){\makebox(0,0)[bl]{1.3.}}

\put(24,130){\makebox(0,0)[bl]{\(0.87\)}}
\put(34,130){\makebox(0,0)[bl]{\(12.8\)}}
\put(44,130){\makebox(0,0)[bl]{\(18.0\)}}

\put(56,130){\makebox(0,0)[bl]{\(1\)}}
\put(66,130){\makebox(0,0)[bl]{\(0\)}}
\put(76,130){\makebox(0,0)[bl]{\(0\)}}


\put(01,126){\makebox(0,0)[bl]{1.4.}}

\put(24,126){\makebox(0,0)[bl]{\(0.25\)}}
\put(34,126){\makebox(0,0)[bl]{\(12.8\)}}
\put(44,126){\makebox(0,0)[bl]{\(18.0\)}}

\put(56,126){\makebox(0,0)[bl]{\(0\)}}
\put(66,126){\makebox(0,0)[bl]{\(0\)}}
\put(76,126){\makebox(0,0)[bl]{\(1\)}}


\put(01,122){\makebox(0,0)[bl]{2.}}

\put(24,122){\makebox(0,0)[bl]{\(1.53\)}}
\put(35,122){\makebox(0,0)[bl]{\(6.4\)}}
\put(45,122){\makebox(0,0)[bl]{\(5.5\)}}

\put(56,122){\makebox(0,0)[bl]{\(2\)}}
\put(66,122){\makebox(0,0)[bl]{\(0\)}}
\put(76,122){\makebox(0,0)[bl]{\(0\)}}


\put(01,118){\makebox(0,0)[bl]{3.}}

\put(24,118){\makebox(0,0)[bl]{\(0.30\)}}
\put(35,118){\makebox(0,0)[bl]{\(9.6\)}}
\put(45,118){\makebox(0,0)[bl]{\(6.4\)}}

\put(56,118){\makebox(0,0)[bl]{\(1\)}}
\put(66,118){\makebox(0,0)[bl]{\(0\)}}
\put(76,118){\makebox(0,0)[bl]{\(0\)}}


\put(01,114){\makebox(0,0)[bl]{4.}}

\put(24,114){\makebox(0,0)[bl]{\(6.80\)}}
\put(34,114){\makebox(0,0)[bl]{\(12.8\)}}
\put(44,114){\makebox(0,0)[bl]{\(18.0\)}}

\put(56,114){\makebox(0,0)[bl]{\(1\)}}
\put(66,114){\makebox(0,0)[bl]{\(0\)}}
\put(76,114){\makebox(0,0)[bl]{\(0\)}}


\put(01,110){\makebox(0,0)[bl]{5.1}}

\put(24,110){\makebox(0,0)[bl]{\(0.00\)}}
\put(35,110){\makebox(0,0)[bl]{\(4.8\)}}
\put(45,110){\makebox(0,0)[bl]{\(5.9\)}}

\put(56,110){\makebox(0,0)[bl]{\(1\)}}
\put(66,110){\makebox(0,0)[bl]{\(0\)}}
\put(76,110){\makebox(0,0)[bl]{\(0\)}}


\put(01,106){\makebox(0,0)[bl]{5.2}}

\put(24,106){\makebox(0,0)[bl]{\(0.16\)}}
\put(35,106){\makebox(0,0)[bl]{\(4.8\)}}
\put(45,106){\makebox(0,0)[bl]{\(5.9\)}}

\put(56,106){\makebox(0,0)[bl]{\(1\)}}
\put(66,106){\makebox(0,0)[bl]{\(0\)}}
\put(76,106){\makebox(0,0)[bl]{\(0\)}}


\put(01,102){\makebox(0,0)[bl]{5.3}}

\put(24,102){\makebox(0,0)[bl]{\(0.94\)}}
\put(35,102){\makebox(0,0)[bl]{\(3.2\)}}
\put(45,102){\makebox(0,0)[bl]{\(3.7\)}}

\put(56,102){\makebox(0,0)[bl]{\(1\)}}
\put(66,102){\makebox(0,0)[bl]{\(0\)}}
\put(76,102){\makebox(0,0)[bl]{\(0\)}}


\put(01,98){\makebox(0,0)[bl]{6.1}}

\put(24,98){\makebox(0,0)[bl]{\(0.10\)}}
\put(35,98){\makebox(0,0)[bl]{\(1.6\)}}
\put(45,98){\makebox(0,0)[bl]{\(1.2\)}}

\put(56,98){\makebox(0,0)[bl]{\(0\)}}
\put(66,98){\makebox(0,0)[bl]{\(0\)}}
\put(76,98){\makebox(0,0)[bl]{\(0\)}}


\put(01,94){\makebox(0,0)[bl]{6.2}}

\put(24,94){\makebox(0,0)[bl]{\(0.50\)}}
\put(35,94){\makebox(0,0)[bl]{\(1.6\)}}
\put(45,94){\makebox(0,0)[bl]{\(1.2\)}}

\put(56,94){\makebox(0,0)[bl]{\(1\)}}
\put(66,94){\makebox(0,0)[bl]{\(0\)}}
\put(76,94){\makebox(0,0)[bl]{\(0\)}}


\put(01,90){\makebox(0,0)[bl]{6.3}}

\put(24,90){\makebox(0,0)[bl]{\(5.60\)}}
\put(35,90){\makebox(0,0)[bl]{\(3.2\)}}
\put(45,90){\makebox(0,0)[bl]{\(4.0\)}}

\put(56,90){\makebox(0,0)[bl]{\(1\)}}
\put(66,90){\makebox(0,0)[bl]{\(0\)}}
\put(76,90){\makebox(0,0)[bl]{\(0\)}}


\put(01,86){\makebox(0,0)[bl]{6.4}}

\put(24,86){\makebox(0,0)[bl]{\(0.81\)}}
\put(35,86){\makebox(0,0)[bl]{\(3.2\)}}
\put(45,86){\makebox(0,0)[bl]{\(3.1\)}}

\put(56,86){\makebox(0,0)[bl]{\(1\)}}
\put(66,86){\makebox(0,0)[bl]{\(0\)}}
\put(76,86){\makebox(0,0)[bl]{\(0\)}}


\put(01,82){\makebox(0,0)[bl]{6.5}}

\put(24,82){\makebox(0,0)[bl]{\(0.35\)}}
\put(35,82){\makebox(0,0)[bl]{\(0.8\)}}
\put(45,82){\makebox(0,0)[bl]{\(1.4\)}}

\put(56,82){\makebox(0,0)[bl]{\(1\)}}
\put(66,82){\makebox(0,0)[bl]{\(1\)}}
\put(76,82){\makebox(0,0)[bl]{\(0\)}}


\put(01,78){\makebox(0,0)[bl]{6.6}}

\put(24,78){\makebox(0,0)[bl]{\(0.35\)}}
\put(35,78){\makebox(0,0)[bl]{\(0.8\)}}
\put(45,78){\makebox(0,0)[bl]{\(1.4\)}}

\put(56,78){\makebox(0,0)[bl]{\(1\)}}
\put(66,78){\makebox(0,0)[bl]{\(1\)}}
\put(76,78){\makebox(0,0)[bl]{\(0\)}}


\put(01,74){\makebox(0,0)[bl]{6.7}}

\put(24,74){\makebox(0,0)[bl]{\(0.20\)}}
\put(35,74){\makebox(0,0)[bl]{\(0.8\)}}
\put(45,74){\makebox(0,0)[bl]{\(1.4\)}}

\put(56,74){\makebox(0,0)[bl]{\(1\)}}
\put(66,74){\makebox(0,0)[bl]{\(1\)}}
\put(76,74){\makebox(0,0)[bl]{\(0\)}}


\put(01,70){\makebox(0,0)[bl]{6.8}}

\put(24,70){\makebox(0,0)[bl]{\(1.50\)}}
\put(34,70){\makebox(0,0)[bl]{\(28.8\)}}
\put(44,70){\makebox(0,0)[bl]{\(48.7\)}}

\put(56,70){\makebox(0,0)[bl]{\(1\)}}
\put(66,70){\makebox(0,0)[bl]{\(1\)}}
\put(76,70){\makebox(0,0)[bl]{\(0\)}}


\put(01,66){\makebox(0,0)[bl]{6.9}}

\put(24,66){\makebox(0,0)[bl]{\(0.70\)}}
\put(35,66){\makebox(0,0)[bl]{\(1.6\)}}
\put(45,66){\makebox(0,0)[bl]{\(2.5\)}}

\put(56,66){\makebox(0,0)[bl]{\(1\)}}
\put(66,66){\makebox(0,0)[bl]{\(1\)}}
\put(76,66){\makebox(0,0)[bl]{\(0\)}}


\put(01,62){\makebox(0,0)[bl]{7.1}}

\put(24,62){\makebox(0,0)[bl]{\(0.00\)}}
\put(35,62){\makebox(0,0)[bl]{\(1.6\)}}
\put(45,62){\makebox(0,0)[bl]{\(1.2\)}}

\put(56,62){\makebox(0,0)[bl]{\(0\)}}
\put(66,62){\makebox(0,0)[bl]{\(0\)}}
\put(76,62){\makebox(0,0)[bl]{\(0\)}}


\put(01,58){\makebox(0,0)[bl]{7.2}}

\put(24,58){\makebox(0,0)[bl]{\(0.35\)}}
\put(35,58){\makebox(0,0)[bl]{\(1.6\)}}
\put(45,58){\makebox(0,0)[bl]{\(1.2\)}}

\put(56,58){\makebox(0,0)[bl]{\(1\)}}
\put(66,58){\makebox(0,0)[bl]{\(0\)}}
\put(76,58){\makebox(0,0)[bl]{\(0\)}}


\put(01,54){\makebox(0,0)[bl]{7.3}}

\put(24,54){\makebox(0,0)[bl]{\(0.00\)}}
\put(35,54){\makebox(0,0)[bl]{\(0.8\)}}
\put(45,54){\makebox(0,0)[bl]{\(1.4\)}}

\put(56,54){\makebox(0,0)[bl]{\(1\)}}
\put(66,54){\makebox(0,0)[bl]{\(0\)}}
\put(76,54){\makebox(0,0)[bl]{\(0\)}}


\put(01,50){\makebox(0,0)[bl]{7.4}}

\put(24,50){\makebox(0,0)[bl]{\(0.20\)}}
\put(35,50){\makebox(0,0)[bl]{\(3.2\)}}
\put(45,50){\makebox(0,0)[bl]{\(3.1\)}}

\put(56,50){\makebox(0,0)[bl]{\(1\)}}
\put(66,50){\makebox(0,0)[bl]{\(0\)}}
\put(76,50){\makebox(0,0)[bl]{\(0\)}}


\put(01,46){\makebox(0,0)[bl]{7.5}}

\put(24,46){\makebox(0,0)[bl]{\(1.50\)}}
\put(35,46){\makebox(0,0)[bl]{\(2.4\)}}
\put(45,46){\makebox(0,0)[bl]{\(2.0\)}}

\put(56,46){\makebox(0,0)[bl]{\(1\)}}
\put(66,46){\makebox(0,0)[bl]{\(2\)}}
\put(76,46){\makebox(0,0)[bl]{\(0\)}}


\put(01,42){\makebox(0,0)[bl]{7.6}}

\put(24,42){\makebox(0,0)[bl]{\(0.00\)}}
\put(35,42){\makebox(0,0)[bl]{\(0.8\)}}
\put(45,42){\makebox(0,0)[bl]{\(1.9\)}}

\put(56,42){\makebox(0,0)[bl]{\(1\)}}
\put(66,42){\makebox(0,0)[bl]{\(0\)}}
\put(76,42){\makebox(0,0)[bl]{\(0\)}}


\put(01,38){\makebox(0,0)[bl]{7.7}}

\put(24,38){\makebox(0,0)[bl]{\(1.50\)}}
\put(35,38){\makebox(0,0)[bl]{\(1.6\)}}
\put(45,38){\makebox(0,0)[bl]{\(2.9\)}}

\put(56,38){\makebox(0,0)[bl]{\(0\)}}
\put(66,38){\makebox(0,0)[bl]{\(0\)}}
\put(76,38){\makebox(0,0)[bl]{\(0\)}}


\put(01,34){\makebox(0,0)[bl]{7.8}}

\put(24,34){\makebox(0,0)[bl]{\(1.40\)}}
\put(35,34){\makebox(0,0)[bl]{\(2.4\)}}
\put(45,34){\makebox(0,0)[bl]{\(2.0\)}}

\put(56,34){\makebox(0,0)[bl]{\(1\)}}
\put(66,34){\makebox(0,0)[bl]{\(0\)}}
\put(76,34){\makebox(0,0)[bl]{\(0\)}}


\put(01,30){\makebox(0,0)[bl]{7.9}}

\put(24,30){\makebox(0,0)[bl]{\(0.70\)}}
\put(35,30){\makebox(0,0)[bl]{\(0.8\)}}
\put(45,30){\makebox(0,0)[bl]{\(1.4\)}}

\put(56,30){\makebox(0,0)[bl]{\(0\)}}
\put(66,30){\makebox(0,0)[bl]{\(0\)}}
\put(76,30){\makebox(0,0)[bl]{\(0\)}}


\put(01,26){\makebox(0,0)[bl]{7.10}}

\put(24,26){\makebox(0,0)[bl]{\(0.20\)}}
\put(35,26){\makebox(0,0)[bl]{\(0.8\)}}
\put(45,26){\makebox(0,0)[bl]{\(1.4\)}}

\put(56,26){\makebox(0,0)[bl]{\(1\)}}
\put(66,26){\makebox(0,0)[bl]{\(0\)}}
\put(76,26){\makebox(0,0)[bl]{\(0\)}}


\put(01,22){\makebox(0,0)[bl]{7.11}}

\put(23.5,22){\makebox(0,0)[bl]{\(70.00\)}}
\put(35,22){\makebox(0,0)[bl]{\(0.8\)}}
\put(45,22){\makebox(0,0)[bl]{\(1.4\)}}

\put(56,22){\makebox(0,0)[bl]{\(2\)}}
\put(66,22){\makebox(0,0)[bl]{\(0\)}}
\put(76,22){\makebox(0,0)[bl]{\(0\)}}


\put(01,18){\makebox(0,0)[bl]{8.}}

\put(24,18){\makebox(0,0)[bl]{\(0.70\)}}
\put(35,18){\makebox(0,0)[bl]{\(3.2\)}}
\put(45,18){\makebox(0,0)[bl]{\(1.4\)}}

\put(56,18){\makebox(0,0)[bl]{\(2\)}}
\put(66,18){\makebox(0,0)[bl]{\(0\)}}
\put(76,18){\makebox(0,0)[bl]{\(0\)}}


\put(01,14){\makebox(0,0)[bl]{8.1}}

\put(24,14){\makebox(0,0)[bl]{\(0.20\)}}
\put(35,14){\makebox(0,0)[bl]{\(3.2\)}}
\put(45,14){\makebox(0,0)[bl]{\(1.4\)}}

\put(56,14){\makebox(0,0)[bl]{\(2\)}}
\put(66,14){\makebox(0,0)[bl]{\(0\)}}
\put(76,14){\makebox(0,0)[bl]{\(0\)}}


\put(01,10){\makebox(0,0)[bl]{8.2}}

\put(24,10){\makebox(0,0)[bl]{\(0.00\)}}
\put(35,10){\makebox(0,0)[bl]{\(3.2\)}}
\put(45,10){\makebox(0,0)[bl]{\(1.4\)}}

\put(56,10){\makebox(0,0)[bl]{\(2\)}}
\put(66,10){\makebox(0,0)[bl]{\(0\)}}
\put(76,10){\makebox(0,0)[bl]{\(0\)}}


\put(01,06){\makebox(0,0)[bl]{8.3}}

\put(24,06){\makebox(0,0)[bl]{\(0.00\)}}
\put(35,06){\makebox(0,0)[bl]{\(3.2\)}}
\put(45,06){\makebox(0,0)[bl]{\(1.4\)}}

\put(56,06){\makebox(0,0)[bl]{\(2\)}}
\put(66,06){\makebox(0,0)[bl]{\(0\)}}
\put(76,06){\makebox(0,0)[bl]{\(0\)}}


\put(01,02){\makebox(0,0)[bl]{8.4}}

\put(24,02){\makebox(0,0)[bl]{\(0.00\)}}
\put(35,02){\makebox(0,0)[bl]{\(3.2\)}}
\put(45,02){\makebox(0,0)[bl]{\(1.4\)}}

\put(56,02){\makebox(0,0)[bl]{\(2\)}}
\put(66,02){\makebox(0,0)[bl]{\(0\)}}
\put(76,02){\makebox(0,0)[bl]{\(0\)}}

\end{picture}
\end{center}

 Multicriteria ranking (sorting) problem is targeted to select
 the most important system component(s) upon criteria as the
 bottleneck(s)
 (e.g., \cite{lev88a,lev12b,levmih88,roy96,zap02}).
 Here, ELECTRE-like technique is used
  (e.g., \cite{lev12b,levmih88,roy96}) based on
  the following criteria weights:~
 \(1.0\) (\(C_{1}\)),
 \(0.3\) (\(C_{2}\)),
 \(0.4\) (\(C_{3}\)),
 \(0.5\) (\(C_{4}\)),
 \(0.2\) (\(C_{5}\)), and
 \(3.0\) (\(C_{6}\)).
 Fig. 4 depicts the results of multicriteria ranking:

 {\it Layer 1} (system bottlenecks):~
  2, 4, 6.3, 6.8, 7.11.

 {\it Layer 2}:~
  1.3, 5.3, 6.4, 7.5, 7.8.

 {\it Layer 3}:~
 1.1, 1.4, 3, 5.1, 5.2, 6.5, 6.6, 6.9, 7.4, 7.7,
 8.1, 8.

  {\it Layer 4}:~
 other components.

\begin{center}
\begin{picture}(114,151)
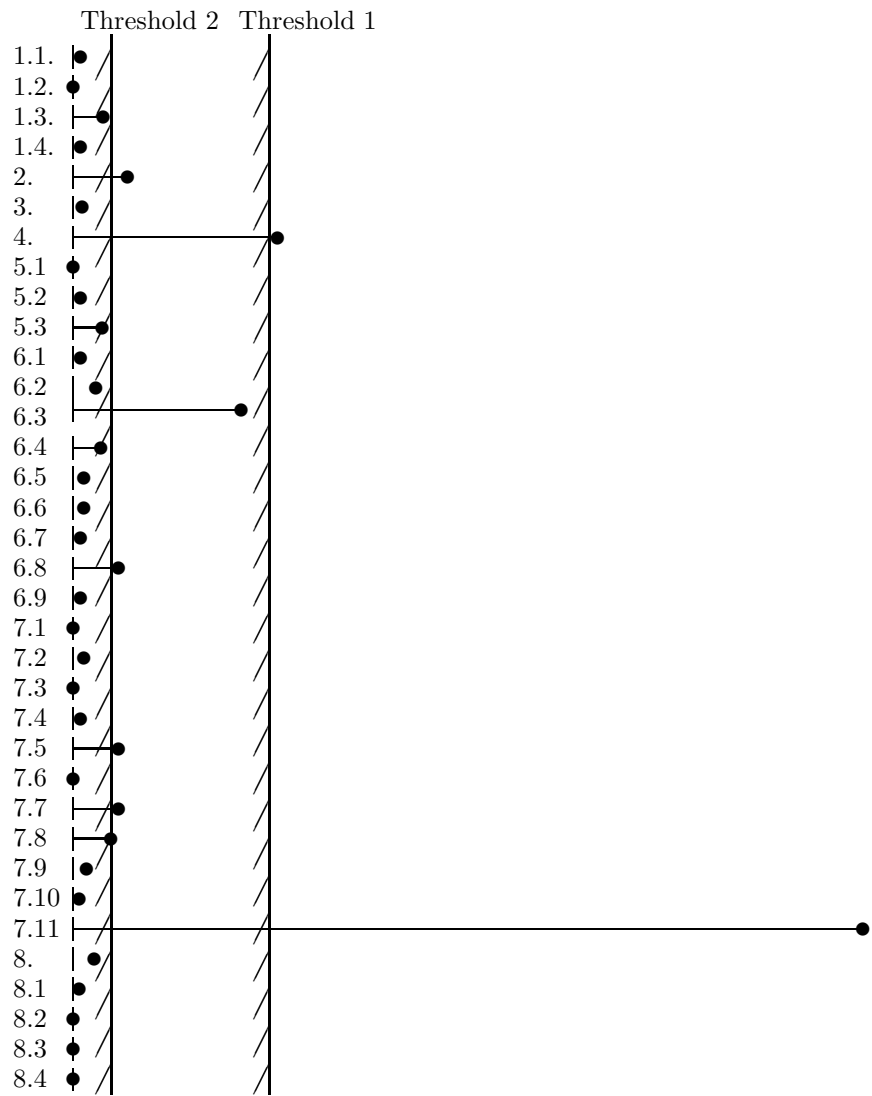
\put(18,00){\makebox(0,0)[bl] {Fig. 3. Pareto chart based solving
procedure}}

\put(09,146.8){\makebox(0,0)[bl] {Threshold \(2\)}}

\put(13,146){\line(0,-1){141}} \put(13.1,146){\line(0,-1){141}}

\put(11,05){\line(1,2){2}} \put(11,10){\line(1,2){2}}
\put(11,15){\line(1,2){2}} \put(11,20){\line(1,2){2}}
\put(11,25){\line(1,2){2}} \put(11,30){\line(1,2){2}}
\put(11,35){\line(1,2){2}} \put(11,40){\line(1,2){2}}
\put(11,45){\line(1,2){2}} \put(11,50){\line(1,2){2}}

\put(11,55){\line(1,2){2}} \put(11,60){\line(1,2){2}}
\put(11,65){\line(1,2){2}} \put(11,70){\line(1,2){2}}
\put(11,75){\line(1,2){2}} \put(11,80){\line(1,2){2}}
\put(11,85){\line(1,2){2}} \put(11,90){\line(1,2){2}}
\put(11,95){\line(1,2){2}} \put(11,100){\line(1,2){2}}

\put(11,105){\line(1,2){2}} \put(11,110){\line(1,2){2}}
\put(11,115){\line(1,2){2}} \put(11,120){\line(1,2){2}}
\put(11,125){\line(1,2){2}} \put(11,130){\line(1,2){2}}
\put(11,135){\line(1,2){2}} \put(11,140){\line(1,2){2}}


\put(30,146.8){\makebox(0,0)[bl] {Threshold \(1\)}}

\put(34,146){\line(0,-1){141}} \put(34.1,146){\line(0,-1){141}}

\put(32,05){\line(1,2){2}} \put(32,10){\line(1,2){2}}
\put(32,15){\line(1,2){2}} \put(32,20){\line(1,2){2}}
\put(32,25){\line(1,2){2}} \put(32,30){\line(1,2){2}}
\put(32,35){\line(1,2){2}} \put(32,40){\line(1,2){2}}
\put(32,45){\line(1,2){2}} \put(32,50){\line(1,2){2}}

\put(32,55){\line(1,2){2}} \put(32,60){\line(1,2){2}}
\put(32,65){\line(1,2){2}} \put(32,70){\line(1,2){2}}
\put(32,75){\line(1,2){2}} \put(32,80){\line(1,2){2}}
\put(32,85){\line(1,2){2}} \put(32,90){\line(1,2){2}}
\put(32,95){\line(1,2){2}} \put(32,100){\line(1,2){2}}

\put(32,105){\line(1,2){2}} \put(32,110){\line(1,2){2}}
\put(32,115){\line(1,2){2}} \put(32,120){\line(1,2){2}}
\put(32,125){\line(1,2){2}} \put(32,130){\line(1,2){2}}
\put(32,135){\line(1,2){2}} \put(32,140){\line(1,2){2}}


\put(00,142){\makebox(0,0)[bl]{1.1.}}


\put(09,143){\circle*{1.7}} \put(08,141.5){\line(0,1){3}}


\put(00,138){\makebox(0,0)[bl]{1.2.}}


\put(08,139){\circle*{1.7}} \put(08,137.5){\line(0,1){3}}


\put(00,134){\makebox(0,0)[bl]{1.3.}}

\put(08,135){\line(1,0){4}} \put(12,135){\circle*{1.7}}
\put(08,133.5){\line(0,1){3}}


\put(00,130){\makebox(0,0)[bl]{1.4.}}


\put(09,131){\circle*{1.7}} \put(08,129.5){\line(0,1){3}}


\put(00,126){\makebox(0,0)[bl]{2.}}

\put(08,127){\line(1,0){7.2}} \put(15.2,127){\circle*{1.7}}
\put(08,125.5){\line(0,1){3}}


\put(00,122){\makebox(0,0)[bl]{3.}}


\put(9.2,123){\circle*{1.7}} \put(08,121.5){\line(0,1){3}}


\put(00,118){\makebox(0,0)[bl]{4.}}

\put(08,119){\line(1,0){27.2}} \put(35.2,119){\circle*{1.7}}
\put(08,117.5){\line(0,1){3}}


\put(00,114){\makebox(0,0)[bl]{5.1}}


\put(08,115){\circle*{1.7}} \put(08,113.5){\line(0,1){3}}


\put(00,110){\makebox(0,0)[bl]{5.2}}


\put(09,111){\circle*{1.7}} \put(08,109.5){\line(0,1){3}}


\put(00,106){\makebox(0,0)[bl]{5.3}}

\put(08,107){\line(1,0){4}} \put(11.88,107){\circle*{1.7}}
\put(08,105.5){\line(0,1){3}}


\put(00,102){\makebox(0,0)[bl]{6.1}}


\put(09,103){\circle*{1.7}} \put(08,101.5){\line(0,1){3}}


\put(00,98){\makebox(0,0)[bl]{6.2}}


\put(11,99){\circle*{1.7}} \put(08,97.5){\line(0,1){3}}


\put(00,94){\makebox(0,0)[bl]{6.3}}

\put(08,96){\line(1,0){22.3}} \put(30.3,96){\circle*{1.7}}
\put(08,94.5){\line(0,1){3}}


\put(00,90){\makebox(0,0)[bl]{6.4}}

\put(08,91){\line(1,0){4}} \put(11.7,91){\circle*{1.7}}
\put(08,89.5){\line(0,1){3}}


\put(00,86){\makebox(0,0)[bl]{6.5}}


\put(9.4,87){\circle*{1.7}} \put(08,85.5){\line(0,1){3}}


\put(00,82){\makebox(0,0)[bl]{6.6}}


\put(9.4,83){\circle*{1.7}} \put(08,81.5){\line(0,1){3}}


\put(00,78){\makebox(0,0)[bl]{6.7}}


\put(09,79){\circle*{1.7}} \put(08,77.5){\line(0,1){3}}


\put(00,74){\makebox(0,0)[bl]{6.8}}

\put(08,75){\line(1,0){6}} \put(14,75){\circle*{1.7}}
\put(08,73.5){\line(0,1){3}}


\put(00,70){\makebox(0,0)[bl]{6.9}}

\put(9,71){\circle*{1.7}}
\put(08,69.5){\line(0,1){3}}


\put(00,66){\makebox(0,0)[bl]{7.1}}


\put(08,67){\circle*{1.7}} \put(08,65.5){\line(0,1){3}}


\put(00,62){\makebox(0,0)[bl]{7.2}}


\put(9.4,63){\circle*{1.7}} \put(08,61.5){\line(0,1){3}}


\put(00,58){\makebox(0,0)[bl]{7.3}}


\put(08,59){\circle*{1.7}} \put(08,57.5){\line(0,1){3}}


\put(00,54){\makebox(0,0)[bl]{7.4}}


\put(09,55){\circle*{1.7}} \put(08,53.5){\line(0,1){3}}


\put(00,50){\makebox(0,0)[bl]{7.5}}

\put(08,51){\line(1,0){6}} \put(14,51){\circle*{1.7}}
\put(08,49.5){\line(0,1){3}}


\put(00,46){\makebox(0,0)[bl]{7.6}}


\put(08,47){\circle*{1.7}} \put(08,45.5){\line(0,1){3}}


\put(00,42){\makebox(0,0)[bl]{7.7}}

\put(08,43){\line(1,0){6}} \put(14,43){\circle*{1.7}}
\put(08,41.5){\line(0,1){3}}


\put(00,38){\makebox(0,0)[bl]{7.8}}

\put(08,39){\line(1,0){5.5}} \put(13,39){\circle*{1.7}}
\put(08,37.5){\line(0,1){3}}


\put(00,34){\makebox(0,0)[bl]{7.9}}


\put(09.8,35){\circle*{1.7}} \put(08,33.5){\line(0,1){3}}


\put(00,30){\makebox(0,0)[bl]{7.10}}


\put(8.8,31){\circle*{1.7}} \put(08,29.5){\line(0,1){3}}


\put(00,26){\makebox(0,0)[bl]{7.11}}

\put(08,27){\line(1,0){105}} \put(113,27){\circle*{1.7}}
\put(08,25.5){\line(0,1){3}}


\put(00,22){\makebox(0,0)[bl]{8.}}


\put(10.8,23){\circle*{1.7}} \put(08,21.5){\line(0,1){3}}


\put(00,18){\makebox(0,0)[bl]{8.1}}


\put(8.8,19){\circle*{1.7}} \put(08,17.5){\line(0,1){3}}


\put(00,14){\makebox(0,0)[bl]{8.2}}


\put(08,15){\circle*{1.7}} \put(08,13.5){\line(0,1){3}}


\put(00,10){\makebox(0,0)[bl]{8.3}}


\put(08,11){\circle*{1.7}} \put(08,09.5){\line(0,1){3}}

\put(00,06){\makebox(0,0)[bl]{8.4}}


\put(08,07){\circle*{1.7}} \put(08,05.5){\line(0,1){3}}

\end{picture}
\end{center}

\begin{center}
\begin{picture}(95,47)
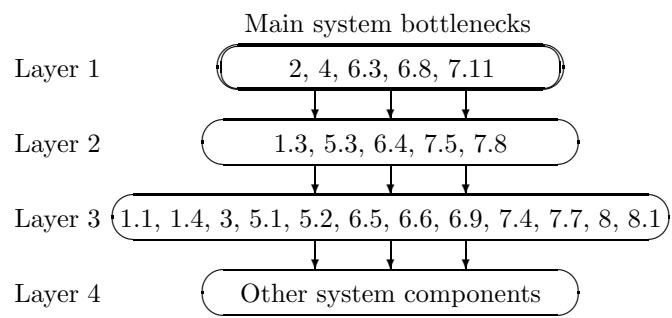
\put(05,00){\makebox(0,0)[bl] {Fig. 4. Results of multicriteria
 ranking (sorting)}}


\put(30.6,43){\makebox(0,0)[bl] {Main system bottlenecks}}

\put(00,37){\makebox(0,0)[bl] {Layer \(1\)}}

\put(50,39){\oval(46,06)} \put(50,39){\oval(45,5.6)}

\put(36.5,37){\makebox(0,0)[bl]{2, 4, 6.3, 6.8, 7.11}}

\put(40,36){\vector(0,-1){4}} \put(50,36){\vector(0,-1){4}}
\put(60,36){\vector(0,-1){4}}


\put(00,27){\makebox(0,0)[bl] {Layer \(2\)}}

\put(50,29){\oval(50,06)}

\put(34.5,27){\makebox(0,0)[bl]{1.3, 5.3, 6.4, 7.5, 7.8}}

\put(40,26){\vector(0,-1){4}} \put(50,26){\vector(0,-1){4}}
\put(60,26){\vector(0,-1){4}}


\put(00,17){\makebox(0,0)[bl] {Layer \(3\)}}

\put(50,19){\oval(74,06)}

\put(14,17){\makebox(0,0)[bl]{1.1, 1.4, 3, 5.1, 5.2, 6.5, 6.6,
 6.9, 7.4, 7.7, 8, 8.1}}

\put(40,16){\vector(0,-1){4}} \put(50,16){\vector(0,-1){4}}
\put(60,16){\vector(0,-1){4}}


\put(00,07){\makebox(0,0)[bl] {Layer \(4\)}}

\put(50,09){\oval(50,06)}

\put(30,07){\makebox(0,0)[bl]{Other system components}}

\end{picture}
\end{center}

\section{Detection of Bottlenecks in Hierarchical Morphological
 Design}

 Hierarchical multicriteria morphological design (HMMD) approach
 for composite (multi-component, modular) systems
 is described in
 \cite{lev98,lev06,lev12morph,lev12a}.
 In HMMD approach,
  the resultant solution is composed from design alternatives
 (DAs) for system parts/components
 while taking into account quality if their interconnection (IC).
%
%
 In the basic version of HMMD, the following ordinal scales are used:
 (1) ordinal scale for quality of system components (or priority)
 (\(\iota = \overline{1,l}\);
      \(1\) corresponds to the best one);
 (2) scale for system quality while taking into account
 system components ordinal estimates
 and ordinal compatibility estimates between the system components
  (\(w=\overline{0,\nu}\); \(\nu\) corresponds to the best level).

 For the system consisting of \(m\) parts/components,
  a discrete space (poset, lattice) of the system quality (excellence) on the basis of the
 following vector is used:
 ~\(N(S)=(w(S);n(S))\),
 ~where \(w(S)\) is the minimum of pairwise compatibility
 between DAs which correspond to different system components,
 ~\(n(S)=(\eta_{1},...,\eta_{r},...,\eta_{k})\),
 ~where ~\(\eta_{r}\) is the number of DAs of the \(r\)th quality in ~\(S\)
 ~(\(\sum^{k}_{r=1} n_{r} = m \)).
 The optimization problem is:
 \[max ~N(S), ~~max ~w(S), ~~w(S) \geq 0.\]
 Let us consider a numerical example (Fig. 5)
 for
 the detection of system bottlenecks in this
 design approach.
 Here, the composite four-component system is:
 \(S = X\star Y\star Z\star H\).
 For each system component, design alternatives (DAs) are depicted in
 in  Fig. 5
 (ordinal estimates of DAs quality as priorities  are presented
 in parentheses, scale \([1,3]\) \(1\) corresponds to the best level of quality).
 Table 3 contains ordinal estimates of compatibility (IC) between DAs
 (scale \([0,3]\)).
%
 Poset-like scales are presented in Fig. 6.

\begin{center}
\begin{picture}(65,34)

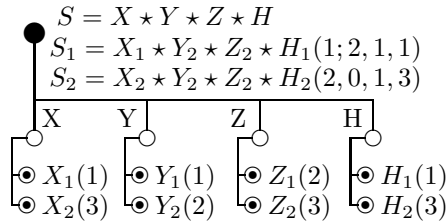
\put(03,00){\makebox(0,0)[bl] {Fig. 5. Four-component system}}

\put(4,05){\makebox(0,8)[bl]{\(X_{2}(3)\)}}
\put(4,09){\makebox(0,8)[bl]{\(X_{1}(1)\)}}

\put(19,05){\makebox(0,8)[bl]{\(Y_{2}(2)\)}}
\put(19,09){\makebox(0,8)[bl]{\(Y_{1}(1)\)}}

\put(34,05){\makebox(0,8)[bl]{\(Z_{2}(3)\)}}
\put(34,09){\makebox(0,8)[bl]{\(Z_{1}(2)\)}}

\put(49,05){\makebox(0,8)[bl]{\(H_{2}(3)\)}}
\put(49,09){\makebox(0,8)[bl]{\(H_{1}(1)\)}}

\put(03,16){\circle{2}} \put(18,16){\circle{2}}
\put(33,16){\circle{2}} \put(48,16){\circle{2}}

\put(00,16){\line(1,0){02}} \put(15,16){\line(1,0){02}}
\put(30,16){\line(1,0){02}} \put(45,16){\line(1,0){02}}

\put(00,16){\line(0,-1){09}} \put(15,16){\line(0,-1){09}}
\put(30,16){\line(0,-1){09}} \put(45,16){\line(0,-1){09}}

\put(45,07){\line(1,0){01}} \put(45,11){\line(1,0){01}}

\put(47,11){\circle{2}} \put(47,11){\circle*{1}}
\put(47,07){\circle{2}} \put(47,07){\circle*{1}}

\put(30,11){\line(1,0){01}} \put(30,07){\line(1,0){01}}

\put(32,11){\circle{2}} \put(32,11){\circle*{1}}
\put(32,07){\circle{2}} \put(32,07){\circle*{1}}

\put(15,07){\line(1,0){01}} \put(15,11){\line(1,0){01}}

\put(17,11){\circle{2}} \put(17,11){\circle*{1}}
\put(17,07){\circle{2}} \put(17,07){\circle*{1}}

\put(0,07){\line(1,0){01}} \put(0,11){\line(1,0){01}}

\put(2,07){\circle{2}} \put(2,11){\circle{2}}
\put(2,07){\circle*{1}} \put(2,11){\circle*{1}}

\put(03,21){\line(0,-1){04}} \put(18,21){\line(0,-1){04}}
\put(33,21){\line(0,-1){04}} \put(48,21){\line(0,-1){04}}

\put(03,21){\line(1,0){45}} \put(03,21){\line(0,1){9}}


\put(03,30){\circle*{2.7}}

\put(04,17.5){\makebox(0,8)[bl]{X}}
\put(14,17.5){\makebox(0,8)[bl]{Y}}
\put(29,17.5){\makebox(0,8)[bl]{Z}}
\put(44,17.5){\makebox(0,8)[bl]{H}}

\put(06,31){\makebox(0,8)[bl]{\(S = X \star Y \star Z \star
 H\)}}

\put(05,26){\makebox(0,8)[bl]{\(S_{1} = X_{1}\star Y_{2}\star
 Z_{2}\star H_{1}(1;2,1,1)\)}}

\put(05,22){\makebox(0,8)[bl] {\(S_{2}=X_{2}\star Y_{2} \star
 Z_{2} \star H_{2}(2,0,1,3)\)}}

\end{picture}
%
\begin{picture}(37,38)

\put(0.7,34){\makebox(0,0)[bl]{Table 3. Compatibility}}

\put(00,0){\line(1,0){37}} \put(00,26){\line(1,0){37}}
\put(00,32){\line(1,0){37}}

\put(00,0){\line(0,1){32}} \put(07,0){\line(0,1){32}}
\put(37,0){\line(0,1){32}}

\put(01,22){\makebox(0,0)[bl]{\(X_{1}\)}}
\put(01,18){\makebox(0,0)[bl]{\(X_{2}\)}}
\put(01,14){\makebox(0,0)[bl]{\(Y_{1}\)}}
\put(01,10){\makebox(0,0)[bl]{\(Y_{2}\)}}
\put(01,06){\makebox(0,0)[bl]{\(Z_{1}\)}}
\put(01,02){\makebox(0,0)[bl]{\(Z_{2}\)}}

\put(12,26){\line(0,1){6}} \put(17,26){\line(0,1){6}}
\put(22,26){\line(0,1){6}} \put(27,26){\line(0,1){6}}
\put(32,26){\line(0,1){6}}

\put(07.4,28){\makebox(0,0)[bl]{\(Y_{1}\)}}
\put(12.4,28){\makebox(0,0)[bl]{\(Y_{2}\)}}
\put(17.4,28){\makebox(0,0)[bl]{\(Z_{1}\)}}
\put(22.4,28){\makebox(0,0)[bl]{\(Z_{2}\)}}
\put(27.4,28){\makebox(0,0)[bl]{\(H_{1}\)}}
\put(32.4,28){\makebox(0,0)[bl]{\(H_{2}\)}}


\put(09,22){\makebox(0,0)[bl]{\(3\)}}
\put(14,22){\makebox(0,0)[bl]{\(2\)}}
\put(19,22){\makebox(0,0)[bl]{\(2\)}}
\put(24,22){\makebox(0,0)[bl]{\(2\)}}
\put(29,22){\makebox(0,0)[bl]{\(1\)}}
\put(34,22){\makebox(0,0)[bl]{\(3\)}}

\put(09,18){\makebox(0,0)[bl]{\(0\)}}
\put(14,18){\makebox(0,0)[bl]{\(3\)}}
\put(19,18){\makebox(0,0)[bl]{\(2\)}}
\put(24,18){\makebox(0,0)[bl]{\(3\)}}
\put(29,18){\makebox(0,0)[bl]{\(1\)}}
\put(34,18){\makebox(0,0)[bl]{\(2\)}}

\put(19,14){\makebox(0,0)[bl]{\(2\)}}
\put(24,14){\makebox(0,0)[bl]{\(2\)}}
\put(29,14){\makebox(0,0)[bl]{\(1\)}}
\put(34,14){\makebox(0,0)[bl]{\(2\)}}

\put(19,10){\makebox(0,0)[bl]{\(2\)}}
\put(24,10){\makebox(0,0)[bl]{\(3\)}}
\put(29,10){\makebox(0,0)[bl]{\(2\)}}
\put(34,10){\makebox(0,0)[bl]{\(3\)}}

\put(29,06){\makebox(0,0)[bl]{\(1\)}}
\put(34,06){\makebox(0,0)[bl]{\(2\)}}

\put(29,02){\makebox(0,0)[bl]{\(3\)}}
\put(34,02){\makebox(0,0)[bl]{\(3\)}}

\end{picture}
\end{center}

\begin{center}
\begin{picture}(60,116)

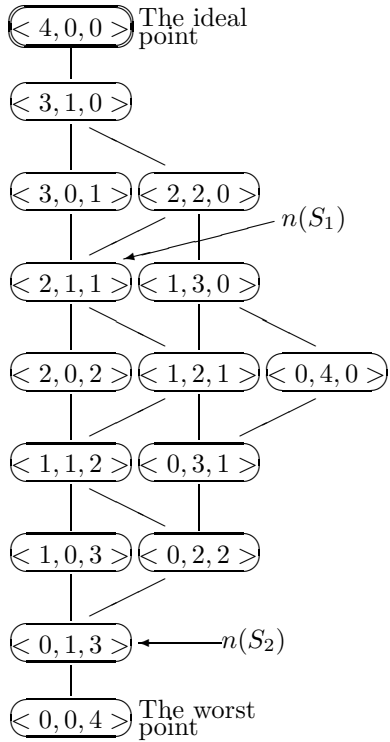
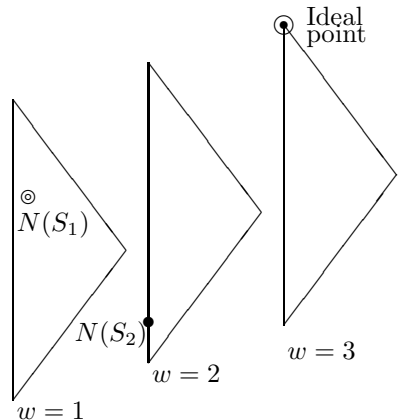
\put(21,00){\makebox(0,0)[bl] {Fig. 6. Poset-like scale for
quality of system \(S\)}}


\put(06,09){\makebox(0,0)[bl] {(a) Poset-like scale}}
\put(11.5,05){\makebox(0,0)[bl]{by elements \(n(S)\)}}

\put(17,111){\makebox(0,0)[bl]{The ideal}}
\put(17,108){\makebox(0,0)[bl]{point}}

\put(00,109){\makebox(0,0)[bl]{\(<4,0,0>\) }}
\put(08,111){\oval(16,5)} \put(08,111){\oval(16.5,5.5)}

\put(08,104){\line(0,1){4}}

\put(00,99){\makebox(0,0)[bl]{\(<3,1,0>\)}}
\put(08,101){\oval(16,5)}

\put(08,92){\line(0,1){6}}

\put(00,87){\makebox(0,0)[bl]{\(<3,0,1>\) }}
\put(08,89){\oval(16,5)}


\put(08,80){\line(0,1){6}}
\put(00,75){\makebox(0,0)[bl]{\(<2,1,1>\)}}
\put(08,77){\oval(16,5)}

\put(08,68){\line(0,1){6}}
\put(00,63){\makebox(0,0)[bl]{\(<2,0,2>\) }}
\put(08,65){\oval(16,5)}
\put(28,27.5){\makebox(0,0)[bl]{\(n(S_{2})\)}}
\put(28,29){\vector(-1,0){11}}



\put(36,83.6){\makebox(0,0)[bl]{\(n(S_{1})\)}}
\put(35,85){\vector(-4,-1){20}}


\put(08,56){\line(0,1){6}}
\put(00,51){\makebox(0,0)[bl]{\(<1,1,2>\) }}
\put(08,53){\oval(16,5)}

\put(08,44){\line(0,1){6}}
\put(00,39){\makebox(0,0)[bl]{\(<1,0,3>\) }}
\put(08,41){\oval(16,5)}

\put(08,32){\line(0,1){6}}
\put(00,27){\makebox(0,0)[bl]{\(<0,1,3>\) }}
\put(08,29){\oval(16,5)}

\put(08,22){\line(0,1){4}}

\put(00,17){\makebox(0,0)[bl]{\(<0,0,4>\) }}
\put(08,19){\oval(16,5)}

\put(17,19){\makebox(0,0)[bl]{The worst}}
\put(17,16){\makebox(0,0)[bl]{point}}


\put(20.5,92.5){\line(-2,1){10}}

\put(20.5,85.5){\line(-2,-1){10}}

\put(17,87){\makebox(0,0)[bl]{\(<2,2,0>\) }}
\put(25,89){\oval(16,5)}

\put(25,80){\line(0,1){6}}
\put(17,75){\makebox(0,0)[bl]{\(<1,3,0>\) }}
\put(25,77){\oval(16,5)}

\put(10.5,73.5){\line(2,-1){10}}

\put(25,68){\line(0,1){6}}
\put(17,63){\makebox(0,0)[bl]{\(<1,2,1>\) }}
\put(25,65){\oval(16,5)}
\put(10.5,56.5){\line(2,1){10}}

\put(25,56){\line(0,1){6}}

\put(17,51){\makebox(0,0)[bl]{\(<0,3,1>\) }}
\put(25,53){\oval(16,5)}
\put(10.5,49.5){\line(2,-1){10}}

\put(25,44){\line(0,1){6}}
\put(17,39){\makebox(0,0)[bl]{\(<0,2,2>\) }}
\put(25,41){\oval(16,5)}
\put(20.5,37.5){\line(-2,-1){10}}


\put(40.5,68.5){\line(-2,1){10}} \put(40.5,61.5){\line(-2,-1){10}}

\put(34,63){\makebox(0,0)[bl]{\(<0,4,0>\) }}
\put(42,65){\oval(16,5)}

\end{picture}
%
\begin{picture}(59,69)


\put(00,09){\makebox(0,0)[bl]{(b) Poset-like scale by elements}}
\put(05.5,05){\makebox(0,0)[bl]{and by compatibility \(N(S)\)}}



\put(00,16){\line(0,1){40}} \put(00,16){\line(3,4){15}}
\put(00,56){\line(3,-4){15}}

\put(18,21){\line(0,1){40}} \put(18,21){\line(3,4){15}}
\put(18,61){\line(3,-4){15}}

\put(36,26){\line(0,1){40}} \put(36,26){\line(3,4){15}}
\put(36,66){\line(3,-4){15}}





\put(18,26.5){\circle*{1.5}}
\put(08.2,23){\makebox(0,0)[bl]{\(N(S_{2})\)}}



\put(02,43){\circle{0.75}} \put(02,43){\circle{1.7}}
\put(00.5,37.6){\makebox(0,0)[bl]{\(N(S_{1})\)}}


\put(36,66){\circle*{1}} \put(36,66){\circle{2.5}}

\put(39,66){\makebox(0,0)[bl]{Ideal}}
\put(39,63){\makebox(0,0)[bl]{point}}


\put(00.5,13.5){\makebox(0,0)[bl]{\(w=1\)}}
\put(18.5,18.5){\makebox(0,0)[bl]{\(w=2\)}}
\put(36.5,21.5){\makebox(0,0)[bl]{\(w=3\)}}

\end{picture}
\end{center}

  Fig. 6a depicts the poset of system quality by components and
 Fig. 6b depicts an integrated poset with compatibility
 (each triangle corresponds to the poset from Fig. 6a).
 Two resultant composite system Pareto-efficient solutions are under examination:
 ~(i)
 \(S_{1} = X_{1}\star Y_{2}\star Z_{2}\star H_{1}\),
 \(N(S_{1}) = (1;2,1,1)\);
 ~(ii) \(S_{2} = X_{2}\star Y_{2}\star Z_{2}\star H_{2}\).
 \(N(S_{2}) = (2;0,1,3)\).

 The system component (DA) or compatibility between a pair of
 DAs can be considered as the system bottleneck(s).
 The following solving schemes (frameworks) can be considered:

~~

 {\it Scheme 1.} Multicriteria ranking of system components (DAs).

 {\it Scheme 2.} Milticriteria ranking of component interconnections.

 {\it Scheme 3.} Joint multicriteria ranking
   of DAs and their interconnections.

 {\it Scheme 4.} Detection of interconnected system component (as a
 composite fault): clique-based fusion.

~~

 Fig. 7 depicts the system solution
 ~\(S_{1} = X_{1}\star Y_{2}\star Z_{2}\star H_{1}\)~
 (including estimates of DAs and their compatibility).
 Table 4 contains six bottlenecks
 (components, their compatibility).
 Evidently, each bottleneck above has to be assessed by some
 criteria in the case of its improvement
 (e.g., possible profit for system quality, required cost).
 Further, it is reasonable to use multicriteria ranking of the
 bottlenecks while taking int account  the above-mentioned
 criteria and to select the most 'prospective'
 bottleneck(s).

\begin{center}
\begin{picture}(65,35)

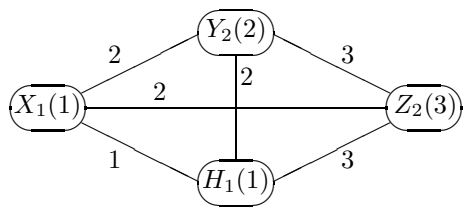
\put(00,00){\makebox(0,0)[bl]{Fig. 7. Concentric presentation of
 solution \(S_{1}\)}}

\put(05,20){\oval(10,6)}
\put(0.5,18.5){\makebox(0,0)[bl]{\(X_{1}(1)\)}}

\put(55,20){\oval(10,6)}
\put(51,18.5){\makebox(0,0)[bl]{\(Z_{2}(3)\)}}

\put(30,30){\oval(10,6)}
\put(26,28.5){\makebox(0,0)[bl]{\(Y_{2}(2)\)}}

\put(30,10){\oval(10,6)}
\put(25.5,08.5){\makebox(0,0)[bl]{\(H_{1}(1)\)}}

\put(19,21){\makebox(0,0)[bl]{\(2\)}}

\put(30.6,23){\makebox(0,0)[bl]{\(2\)}}

\put(13,26){\makebox(0,0)[bl]{\(2\)}}

\put(44,26){\makebox(0,0)[bl]{\(3\)}}

\put(13,12){\makebox(0,0)[bl]{\(1\)}}

\put(44,12){\makebox(0,0)[bl]{\(3\)}}


\put(10,20){\line(1,0){40}} \put(30,13){\line(0,1){14}}

\put(09.5,22){\line(2,1){15.4}} \put(50.5,22){\line(-2,1){15.4}}

\put(09.5,18){\line(2,-1){15.4}} \put(50.5,18){\line(-2,-1){15.4}}

\end{picture}
\end{center}

\begin{center}
\begin{picture}(91,45)

\put(001,41){\makebox(0,0)[bl]{Table 4. Bottlenecks,  possible
improvement actions for \(S_{1}\)}}

\put(00,00){\line(1,0){91}} \put(00,27){\line(1,0){91}}
\put(46,33){\line(1,0){30}} \put(00,39){\line(1,0){91}}

\put(00,0){\line(0,1){39}} \put(46,0){\line(0,1){39}}
\put(58,0){\line(0,1){33}} \put(76,0){\line(0,1){39}}
\put(91,0){\line(0,1){39}}

\put(01,34.4){\makebox(0,0)[bl]{Composite DAs}}
\put(52,35){\makebox(0,0)[bl]{Bottlenecks}}
\put(48,29){\makebox(0,0)[bl]{DAs}}
\put(65,29){\makebox(0,0)[bl]{IC}}
\put(77,35){\makebox(0,0)[bl]{Actions}}
\put(80,31){\makebox(0,0)[bl]{\(w/\iota\)}}


\put(01,22){\makebox(0,0)[bl]{1. \(S_{1} = X_{1}\star Y_{2}\star
Z_{2}\star H_{1}\)}}

\put(01,18){\makebox(0,0)[bl]{2. \(S_{1} = X_{1}\star Y_{2}\star
Z_{2}\star H_{1}\)}}

\put(01,14){\makebox(0,0)[bl]{3. \(S_{1} = X_{1}\star Y_{2}\star
Z_{2}\star H_{1}\)}}

\put(01,10){\makebox(0,0)[bl]{4. \(S_{1} = X_{1}\star Y_{2}\star
Z_{2}\star H_{1}\)}}

\put(01,06){\makebox(0,0)[bl]{5. \(S_{1} = X_{1}\star Y_{2}\star
Z_{2}\star H_{1}\)}}

\put(01,02){\makebox(0,0)[bl]{6. \(S_{1} = X_{1}\star Y_{2}\star
Z_{2}\star H_{1}\)}}

\put(79,22){\makebox(0,0)[bl]{\(2 \Rightarrow 1\)}}
\put(79,18){\makebox(0,0)[bl]{\(2 \Rightarrow 1\)}}
\put(79,14){\makebox(0,0)[bl]{\(2 \Rightarrow 3\)}}
\put(79,10){\makebox(0,0)[bl]{\(2 \Rightarrow 3\)}}
\put(79,06){\makebox(0,0)[bl]{\(1 \Rightarrow 3\)}}
\put(79,02){\makebox(0,0)[bl]{\(2 \Rightarrow 3\)}}

\put(49.5,22){\makebox(0,0)[bl]{\(Y_{1}\)}}
\put(49.5,18){\makebox(0,0)[bl]{\(Z_{2}\)}}

\put(60.5,14){\makebox(0,0)[bl]{\((X_{1},Y_{2})\)}}
\put(60.5,10){\makebox(0,0)[bl]{\((X_{1},Z_{2})\)}}
\put(60.5,06){\makebox(0,0)[bl]{\((X_{1},H_{1})\)}}
\put(60.5,02){\makebox(0,0)[bl]{\((Y_{2},H_{1})\)}}

\end{picture}
\end{center}

 The detection of the system bottleneck
 as a group of interconnected system components
 can be considered as revelation of
 a set of low quality components
 which are connected at the high level compatibility.
 This situation corresponds to a new type of a composite system fault
 which was suggested in \cite{lev12clique,levlast06}.
 Here, some weak system faults are interconnected (at the high
 level) and this combination can lead to a significant composite system
 fault.
 In our case,
 the composite bottleneck (as the composite fault) corresponds to
 the combination of low quality components with high-level
 component compatibility.
 Thus, the following two-criteria optimization problem can be examined:~

~~

   \(min ~n(B),~~  max ~w(B)\);  ~\(B\) is a subsolution of a system solution for \(S\).

~~

 Fig. 8 illustrates a composite solution
 ~\(S_{2}= X_{2} \star Y_{2} \star Z_{2} \star H_{1}\),
 \(N(S_{2}) = (2;0,1,3)\)
 (from example in Fig. 5).
 For this four-component solution,
  it is possible to examine four three-component subsystems:~
 ~\(B_{1} = X_{2} \star Y_{2} \star Z_{2}\), \(N(B_{1}) = (2;0,1,2)\);
 ~\(B_{2} = X_{2} \star Z_{2} \star H_{2}\), \(N(B_{2}) = (2;0,0,3)\);
 ~\(B_{3} = X_{2} \star Y_{2} \star H_{2}\), \(N(B_{3}) = (2;0,2,1)\);
 ~\(B_{4} = Y_{2} \star Z_{2} \star H_{2}\), \(N(B_{4}) = (3;0,2,1)\).
 Two Pareto-efficient subsystems as composite bottlenecks are (Fig. 9):
 ~\(B_{2}\) and ~\(B_{4}\).

\begin{center}
\begin{picture}(65,35)

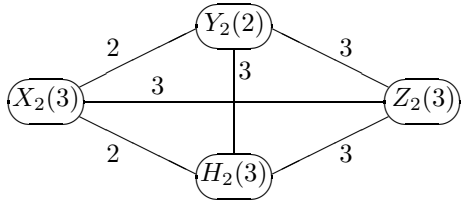
\put(00,00){\makebox(0,0)[bl]{Fig. 8. Concentric presentation of
 solution \(S_{2}\)}}

\put(05,20){\oval(10,6)}
\put(0.5,18.5){\makebox(0,0)[bl]{\(X_{2}(3)\)}}

\put(55,20){\oval(10,6)}
\put(51,18.5){\makebox(0,0)[bl]{\(Z_{2}(3)\)}}

\put(30,30){\oval(10,6)}
\put(26,28.5){\makebox(0,0)[bl]{\(Y_{2}(2)\)}}

\put(30,10){\oval(10,6)}
\put(25.5,08.5){\makebox(0,0)[bl]{\(H_{2}(3)\)}}

\put(19,21){\makebox(0,0)[bl]{\(3\)}}

\put(30.6,23){\makebox(0,0)[bl]{\(3\)}}

\put(13,26){\makebox(0,0)[bl]{\(2\)}}

\put(44,26){\makebox(0,0)[bl]{\(3\)}}

\put(13,12){\makebox(0,0)[bl]{\(2\)}}

\put(44,12){\makebox(0,0)[bl]{\(3\)}}


\put(10,20){\line(1,0){40}} \put(30,13){\line(0,1){14}}

\put(09.5,22){\line(2,1){15.4}} \put(50.5,22){\line(-2,1){15.4}}

\put(09.5,18){\line(2,-1){15.4}} \put(50.5,18){\line(-2,-1){15.4}}

\end{picture}
\end{center}

\begin{center}
\begin{picture}(55,87)

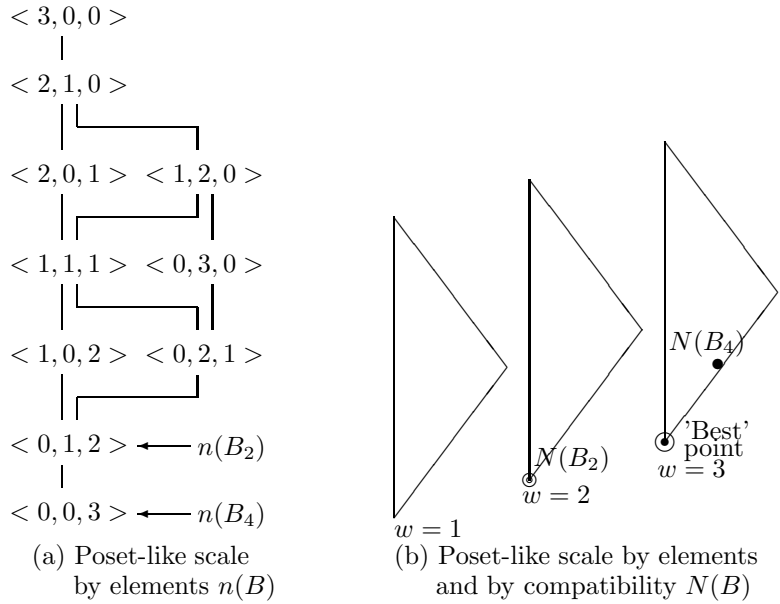
\put(17,00){\makebox(0,0)[bl] {Fig. 9. Poset-like scale for
quality of subsystem \(B\)}}

\put(08,09){\makebox(0,0)[bl] {(a) Poset-like scale}}
\put(13.5,05){\makebox(0,0)[bl]{by elements \(n(B)\)}}

\put(05,81){\makebox(0,0)[bl]{\(<3,0,0>\) }}

\put(12,77){\line(0,1){3}}
\put(05,72){\makebox(0,0)[bl]{\(<2,1,0>\)}}

\put(12,65){\line(0,1){6}}
\put(05,60){\makebox(0,0)[bl]{\(<2,0,1>\) }}

\put(12,53){\line(0,1){6}}
\put(05,48){\makebox(0,0)[bl]{\(<1,1,1>\) }}

\put(12,41){\line(0,1){6}}
\put(05,36){\makebox(0,0)[bl]{\(<1,0,2>\) }}


\put(12,29){\line(0,1){6}}
\put(05,24){\makebox(0,0)[bl]{\(<0,1,2>\) }}

\put(30,23.5){\makebox(0,0)[bl]{\(n(B_{2})\)}}
\put(29,25.5){\vector(-1,0){07}}

\put(30,14.5){\makebox(0,0)[bl]{\(n(B_{4})\)}}
\put(29,16.5){\vector(-1,0){07}}

\put(12,20){\line(0,1){3}}

\put(05,15){\makebox(0,0)[bl]{\(<0,0,3>\) }}


\put(14,68){\line(0,1){3}} \put(30,68){\line(-1,0){16}}
\put(30,65){\line(0,1){3}}

\put(23,60){\makebox(0,0)[bl]{\(<1,2,0>\) }}

\put(30,59){\line(0,-1){3}} \put(30,56){\line(-1,0){16}}
\put(14,56){\line(0,-1){3}}
\put(32,53){\line(0,1){6}}
\put(23,48){\makebox(0,0)[bl]{\(<0,3,0>\) }}

\put(14,44){\line(0,1){3}} \put(30,44){\line(-1,0){16}}
\put(30,41){\line(0,1){3}}

\put(32,41){\line(0,1){6}}
\put(23,36){\makebox(0,0)[bl]{\(<0,2,1>\) }}

\put(30,35){\line(0,-1){3}} \put(30,32){\line(-1,0){16}}
\put(14,32){\line(0,-1){3}}

\end{picture}
%
\begin{picture}(59,69)

\put(00,09){\makebox(0,0)[bl]{(b) Poset-like scale by elements}}
\put(05.5,05){\makebox(0,0)[bl]{and by compatibility \(N(B)\)}}

\put(00,16){\line(0,1){40}} \put(00,16){\line(3,4){15}}
\put(00,56){\line(3,-4){15}}

\put(18,21){\line(0,1){40}} \put(18,21){\line(3,4){15}}
\put(18,61){\line(3,-4){15}}

\put(36,26){\line(0,1){40}} \put(36,26){\line(3,4){15}}
\put(36,66){\line(3,-4){15}}


\put(43,36.5){\circle*{1.5}}
\put(36.5,37.5){\makebox(0,0)[bl]{\(N(B_{4})\)}}


\put(18,21){\circle{0.75}} \put(18,21){\circle{1.7}}
\put(18.5,22){\makebox(0,0)[bl]{\(N(B_{2})\)}}


\put(36,26){\circle*{1}} \put(36,26){\circle{2.5}}

\put(38.5,26.5){\makebox(0,0)[bl]{'Best'}}
\put(38.5,23.5){\makebox(0,0)[bl]{point}}

\put(00,13.5){\makebox(0,0)[bl]{\(w=1\)}}
\put(17,18){\makebox(0,0)[bl]{\(w=2\)}}
\put(35,21.5){\makebox(0,0)[bl]{\(w=3\)}}

\end{picture}
\end{center}


\section{Critical Elements in Multilayer Structures/Networks}

 Generally, it is reasonable to examine multi-layer structures/networks
 (Fig. 10) (e.g., \cite{lev12hier}).

\begin{center}
\begin{picture}(62,50)

\put(02.5,00){\makebox(0,0)[bl]{Fig. 10. Multilayer structure
\cite{lev12hier,lev13imp}}}



\put(48,42){\makebox(0,0)[bl]{Top}}
\put(48,39){\makebox(0,0)[bl]{layer}}

\put(02.5,41.2){\makebox(0,0)[bl]{Layer nodes \(L'_{t} =
\{1,...,m'_{t}\}\)}}

\put(24,42){\oval(46,8)}

\put(09,40){\circle*{1.8}} \put(13,40){\circle*{1.8}}
\put(20.5,39.5){\makebox(0,0)[bl]{{\bf .~.~.}}}

\put(35,40){\circle*{1.8}}  \put(39,40){\circle*{1.8}}


\put(09,40){\vector(0,-1){06}} \put(09,40){\vector(-1,-2){03}}
\put(13,40){\vector(0,-1){06}}

\put(35,40){\vector(0,-1){06}} \put(35,40){\vector(-1,-2){03}}

\put(39,40){\vector(0,-1){06}} \put(39,40){\vector(-1,-2){03}}
\put(39,40){\vector(1,-2){03}}

\put(8,31){\makebox(0,0)[bl]{Connection for nodes}}
\put(10,28){\makebox(0,0)[bl]{of neighbor layers}}


\put(48,25){\makebox(0,0)[bl]{Inter-}}
\put(48,22){\makebox(0,0)[bl]{mediate}}
\put(48,19){\makebox(0,0)[bl]{layer}}

\put(03,23.2){\makebox(0,0)[bl]{Layer nodes \(L'_{i} =
\{1,...,m'_{i}\}\)}}

\put(24,24){\oval(46,8)}

\put(07,22){\circle*{1.5}} \put(12,22){\circle*{1.5}}
\put(20.5,21.5){\makebox(0,0)[bl]{{\bf .~.~.}}}

\put(37,22){\circle*{1.5}}  \put(42,22){\circle*{1.5}}


\put(07,22){\vector(0,-1){06}} \put(07,22){\vector(-1,-2){03}}
\put(07,22){\vector(1,-2){03}}

\put(12,22){\vector(0,-1){06}} \put(12,22){\vector(1,-2){03}}

\put(37,22){\vector(0,-1){06}} \put(37,22){\vector(-1,-2){03}}

\put(42,22){\vector(0,-1){06}} \put(42,22){\vector(-1,-2){03}}
\put(42,22){\vector(1,-2){03}}





\put(49.6,10){\makebox(0,0)[bl]{Bottom}}
\put(49.6,07){\makebox(0,0)[bl]{layer}}

\put(24,10){\oval(48,8)}

\put(02,09.2){\makebox(0,0)[bl]{Layer nodes \(L'_{b} =
\{1,...,m'_{b}\}\)}}

\put(03,08){\circle*{1}} \put(08,08){\circle*{1}}
\put(20.5,07.5){\makebox(0,0)[bl]{{\bf .~.~.}}}

\put(38,08){\circle*{1}}  \put(43,08){\circle*{1}}

\end{picture}
\end{center}
 Here, the following kinds of problems
 for detection of system bottlenecks can be examined:

 {\it Kind I:}~ for structure/network layer:

  (i) detection of critical nodes in networks
  (e.g., maximum leaf spanning tree problem,
  connecting connected dominating sets problem),

  (ii) detection of group of critical network nodes,

 (iii) detection of group of critical interconnected network nodes,
 and

 (iv) detection of low quality layer topology.

  {\it Kind II:}~   for neighbor layers:
  detection of critical
  connection between nodes of neighbor layers.

 {\it Kind III:}~  for multi-layers:
  detection of wrong or low quality assignment of nodes into
  structure/network layers.

 Let us consider some of the problems above
 for the structure layer level.

  First, detection of critical node(s) in the structure/network layer
  may be based on the methods which were describes
 in previous sections
 (e.g., Pareto chart based method, multicriteria
 analysis/ranking,
 detection of interconnected nodes as clique fusion).
 Second, three well-known combinatorial
 optimization problems can be considered.
%
 Fig. 11 illustrates this type of combinatorial problems:
 (a) {\it maximum leaf/terminal nodes problem},
 (b) {\it minimum internal nodes problem}, and
 (c) {\it hierarchical two-level network design problem}.
 Here,
 the set of internal structure/network nodes
 can be considered as some crucial nodes
 (e.g., for improvement, for testing) or 'bottlenecks'.

\begin{center}
\begin{picture}(35,21)
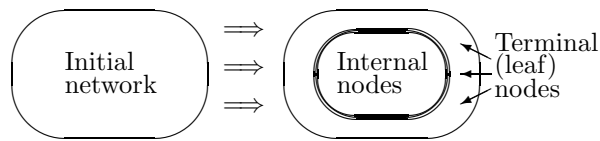
\put(0.5,00){\makebox(0,0)[bl]{Fig. 11. Maximum
 leaf/ minimum internal nodes}}

\put(07,14.5){\makebox(0,0)[bl]{Initial}}
\put(07,11.5){\makebox(0,0)[bl]{network}}

\put(13,14){\oval(26,17)}


\put(28,19){\makebox(0,0)[bl]{\(\Longrightarrow \)}}
\put(28,14){\makebox(0,0)[bl]{\(\Longrightarrow \)}}
\put(28,09){\makebox(0,0)[bl]{\(\Longrightarrow \)}}


\end{picture}
%
\begin{picture}(43,27)


\put(13,14){\oval(26,17)}

\put(28,17){\makebox(0,0)[bl]{Terminal}}
\put(28,13.3){\makebox(0,0)[bl]{(leaf)}}
\put(28,11){\makebox(0,0)[bl]{nodes}}

\put(27.5,16){\vector(-2,1){4}} \put(27.5,14){\vector(-1,0){4}}
\put(27.5,12){\vector(-2,-1){4}}

\put(13,14){\oval(18,12)} \put(13,14){\oval(17.5,11.5)}
\put(13,14){\oval(17,11)}

\put(07,14.5){\makebox(0,0)[bl]{Internal}}
\put(07,11.5){\makebox(0,0)[bl]{nodes}}

\end{picture}
\end{center}


 The ``maximum leaf spanning tree'' problem is the following
 (e.g.,
 \cite{alon09,caro00,gar79}):

~~~

  Find a spanning tree of an input graph
 so that the number of the tree leafs is maximal.

~~

 Generally,
 the spanning tree of a graph contains the following types of nodes:
 (a) root,
 (b) internal nodes
  (the internal nodes may be considered as a virtual
 ``bus'' in networking),
  and
 (c) leaf nodes.
 Thus, the problem consists in maximizing the number of leaf nodes
 or
 minimizing the number of internal nodes.
 The problem is one of the basic NP-hard problems \cite{gar79}.
%

%
%
%
%
%
%
%
%
%
%

%
 In sense of exact algorithms, this problem is equivalent to
 ``connected dominating set'' problem (NP-hard)
 (e.g., \cite{blum05,caro00,gar79}):

~~

 Find a minimum set of vertices \(D \subseteq A\)
 of  input graph \(G = (A,E)\)
 that the induced by \(D\) subgraph \(G'=(D,E')\) (\( E' \subseteq E \))
 is connected dominated set and \(D\) is a dominating set of \(G\).

~~

  A recent survey on the connected dominating set problem is presented in \cite{blum05}.

%
%
%
%
%


 The basic hierarchical two-level network design problem is
 (e.g., \cite{bala94,current86,pirkul91}):

~~

 Find  a minimum cost two-level spanning network,
 consisting of two parts:
 (i) main (internal) path (or several paths, tree, ring)
 (ii) secondary trees.
%

~~

 Thus, the initial network is divided into two parts:

 (a) main part (i.e., the higher level part):
  a path (or several paths, tree, ring)
 composed of primary arcs,
 which visits some of the nodes of the network
 (i.e., primary nodes);

 (b) secondary part (i.e., secondary nodes, secondary trees):
 the part is composed of one or more
 trees whose arcs, termed secondary,
 are less expensive to build than the primary arcs.

 Here,
 each arc has a cost (\(d_{ij},~ \forall i,j \in A\), \(A\) is the set of
 nodes).
 The total cost of the selected arcs in the spanning structure
 is used as the minimized objective function.
 The problem is formulated as combinatorial optimization model
 (e.g., \cite{current86}), it is NP-hard \cite{bala94}.
%
%
%
%
%
%

 Evidently, similar problems can be considered for detection of
  critical arcs in networks.

 In the above-mentioned problems {\it kind II} and {\it kind III},
 the solution consists in  assignment of elements into positions
 (i.e., assignment/allocation problems).
 Here, new advanced combinatorial problem statements
 are required for the detection of low quality assignment(s) in the
 existing solution(s).
 Note,
 usage of HMMD approach to an extended assignment problem
 has been suggested in \cite{lev98,lev09}.
 Thus, detection of bottlenecks in hierarchical morphological
 design, described in previous section,
 can be used for the assignment/allocation problems as well.

 Detection of low quality
 network topology requires special additional study.
  The augmentation problem
  (e.g., \cite{esw76})
   can be considered as
 a version of this approach.

\section{Predictive Detection of System Bottlenecks}

 A predictive detection of system bottleneck(s)
 can be considered as the following (Fig. 12):


 {\it Step 1.} Study of existing changes and/or future changes
  of systems parameters and/or
 system structure
 (i.e., parameters for system components,
 parameters for system structure).

 {\it Step 2.} Analysis of system evolution
 (i.e., the corresponding trajectories for system,
 system parameters).

 {\it Step 3.} Forecasting  of the system parameters
 to build the system forecast.

 {\it Step 4.} Detection of the system bottleneck(s)
 on the basis of the future system parameters
 (i.e., system forecast, system parameters forecasts).

\begin{center}
\begin{picture}(81,41)
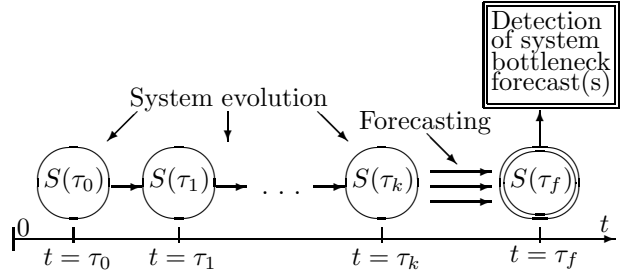
\put(00,00){\makebox(0,0)[bl] {Fig. 12. Predictive detection of
 system bottleneck(s)}}

\put(0.5,9.5){\makebox(0,8)[bl]{\(0\)}}
\put(00,7.5){\line(0,1){3}}

\put(00,9){\vector(1,0){80}} \put(78,10){\makebox(0,8)[bl]{\(t\)}}

\put(4,5){\makebox(0,8)[bl]{\(t=\tau_{0}\)}}
\put(18,5){\makebox(0,8)[bl]{\(t=\tau_{1}\)}}
\put(45,5){\makebox(0,8)[bl]{\(t=\tau_{k}\)}}

\put(66,5){\makebox(0,8)[bl]{\(t=\tau_{f}\)}}

\put(08,8.5){\line(0,1){2}} \put(22,8.5){\line(0,1){2}}
\put(49,8.5){\line(0,1){2}} \put(70,8.5){\line(0,1){2}}


\put(4,15){\makebox(0,8)[bl]{\(S(\tau_{0})\)}}

\put(08,16.5){\oval(09.5,9.5)} \put(13,16){\vector(1,0){4}}


\put(18,15){\makebox(0,8)[bl]{\(S(\tau_{1})\)}}

\put(22,16.5){\oval(09.5,9.5)}

\put(27,16){\vector(1,0){4}}

\put(33,15){\makebox(0,8)[bl]{.~.~.}}

\put(40,16){\vector(1,0){4}}


\put(45,15){\makebox(0,8)[bl]{\(S(\tau_{k})\)}}

\put(49,16.5){\oval(09.5,9.5)}

\put(55.5,18){\vector(1,0){08}} \put(55.5,16){\vector(1,0){08}}
\put(55.5,14){\vector(1,0){08}}


\put(15.5,26){\makebox(0,8)[bl]{System evolution}}

\put(16.5,26){\vector(-1,-1){4}} \put(28.5,26){\vector(0,-1){4}}
\put(40.5,26){\vector(1,-1){4}}


\put(46,23){\makebox(0,8)[bl]{Forecasting}}
\put(55,23){\vector(1,-1){4}}


\put(66,15){\makebox(0,8)[bl]{\(S(\tau_{f})\)}}

\put(70,16.5){\oval(10.5,9.5)} \put(70,16.5){\oval(09.5,8.5)}



\put(63,27){\line(1,0){17}}\put(63,40){\line(1,0){17}}
\put(63,27){\line(0,1){13}}\put(80,27){\line(0,1){13}}

\put(62.5,26.5){\line(1,0){18}}\put(62.5,40.5){\line(1,0){18}}
\put(62.5,26.5){\line(0,1){14}}\put(80.5,26.5){\line(0,1){14}}

\put(63.5,37){\makebox(0,8)[bl]{Detection}}
\put(63.5,33.7){\makebox(0,8)[bl]{of system}}
\put(63.5,31.5){\makebox(0,8)[bl]{bottleneck}}
\put(63.5,28){\makebox(0,8)[bl]{forecast(s)}}

\put(70,21.5){\vector(0,1){4.6}}

\end{picture}
\end{center}

 Evidently,
 the same system objects can be under examination:
 system component(s),
 group of interconnected system components,
 system structure.
%

  In the case of network-like system,
 the pointed out predictive detection problems can be  complicated.

\subsection{Predictive Detection of System Components}

 The predictive detection of system
 bottlenecks as system component(s)
 can be based on the same methods
 (i.e., Pareto chart method, multicriteria ranking).
 In this case,
 system parameters forecasts are used as the initial information.
 In the considered example for aggregate
 (Fig. 2, Table 2, Fig. 3, Fig. 4),
 forecasts of the data from Table 2 have to be used.

\subsection{Predictive Detection of Interconnected System Components}

 The predictive detection of bottlenecks in hierarchical
 morphological design can be considered analogically
 (i.e., analysis of the system evolution, computing a system forecast,
 detection of system bottleneck(s) via the methods above
 for the system forecast).

 Let us consider a simplified example for
 detection of a composite bottleneck (as a subsystem)
 for four-component system
 \(S = X\star Y \star Z \star H\)
  from Fig. 5.
 Fig. 13 depicts an illustrative numerical example for evolution
 and forecasting of solution
 ~\(S_{2} = X_{2} \star Y_{2} \star Z_{2} \star H_{2}\)
 ~(\(N(S_{2})=2;0,1,3\)).
 Here, the following time axe is considered:
 basic time point \(t=\tau_{0}\),
 next time point (evolution) \(t=\tau_{1}\),
 forecast time point  \(t=\tau_{2}\)
 (i.e., \(t=\tau_{f}\)).

\begin{center}
\begin{picture}(40.5,41.5)

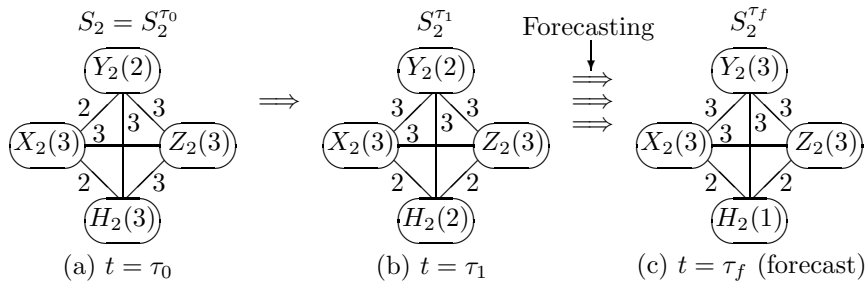
\put(021,00){\makebox(0,0)[bl]{Fig. 13. Evolution of solution
\(S_{2}\) and forecast}}

\put(09,37.5){\makebox(0,0)[bl]{\(S_{2}=S_{2}^{\tau_{0}}\)}}

\put(07,05){\makebox(0,0)[bl]{(a) \(t=\tau_{0}\)}}


\put(05,23){\oval(10,6)}
\put(0.5,21.5){\makebox(0,0)[bl]{\(X_{2}(3)\)}}

\put(25,23){\oval(10,6)}
\put(21,21.5){\makebox(0,0)[bl]{\(Z_{2}(3)\)}}

\put(15,33){\oval(10,6)}
\put(11,31.5){\makebox(0,0)[bl]{\(Y_{2}(2)\)}}

\put(15,13){\oval(10,6)}
\put(10.5,11.5){\makebox(0,0)[bl]{\(H_{2}(3)\)}}

\put(11,23.5){\makebox(0,0)[bl]{\(3\)}}

\put(15.6,25){\makebox(0,0)[bl]{\(3\)}}

\put(09,27){\makebox(0,0)[bl]{\(2\)}}

\put(19,27){\makebox(0,0)[bl]{\(3\)}}

\put(09,17){\makebox(0,0)[bl]{\(2\)}}

\put(19,17){\makebox(0,0)[bl]{\(3\)}}


\put(33,28){\makebox(0,0)[bl]{\(\Longrightarrow\)}}


\put(10,23){\line(1,0){10}} \put(15,16){\line(0,1){14}}

\put(09.5,25){\line(1,1){05}} \put(20.5,25){\line(-1,1){05}}

\put(09.5,21){\line(1,-1){05}} \put(20.5,21){\line(-1,-1){05}}

\end{picture}
%
\begin{picture}(40.5,41)

\put(12.4,37.5){\makebox(0,0)[bl]{\(S_{2}^{\tau_{1}}\)}}

\put(07,05){\makebox(0,0)[bl]{(b) \(t=\tau_{1}\)}}


\put(05,23){\oval(10,6)}
\put(0.5,21.5){\makebox(0,0)[bl]{\(X_{2}(3)\)}}

\put(25,23){\oval(10,6)}
\put(21,21.5){\makebox(0,0)[bl]{\(Z_{2}(3)\)}}

\put(15,33){\oval(10,6)}
\put(11,31.5){\makebox(0,0)[bl]{\(Y_{2}(2)\)}}

\put(15,13){\oval(10,6)}
\put(10.5,11.5){\makebox(0,0)[bl]{\(H_{2}(2)\)}}

\put(11,23.5){\makebox(0,0)[bl]{\(3\)}}

\put(15.6,25){\makebox(0,0)[bl]{\(3\)}}

\put(09,27){\makebox(0,0)[bl]{\(3\)}}

\put(19,27){\makebox(0,0)[bl]{\(3\)}}

\put(09,17){\makebox(0,0)[bl]{\(2\)}}

\put(19,17){\makebox(0,0)[bl]{\(2\)}}


\put(26.5,37){\makebox(0,0)[bl]{Forecasting}}
\put(35.6,37){\vector(0,-1){04}}

\put(33,25){\makebox(0,0)[bl]{\(\Longrightarrow\)}}
\put(33,31){\makebox(0,0)[bl]{\(\Longrightarrow\)}}
\put(33,28){\makebox(0,0)[bl]{\(\Longrightarrow\)}}


\put(10,23){\line(1,0){10}} \put(15,16){\line(0,1){14}}

\put(09.5,25){\line(1,1){05}} \put(20.5,25){\line(-1,1){05}}

\put(09.5,21){\line(1,-1){05}} \put(20.5,21){\line(-1,-1){05}}

\end{picture}
%
\begin{picture}(35,41)

\put(12.4,37.5){\makebox(0,0)[bl]{\(S_{2}^{\tau_{f}}\)}}

\put(00,05){\makebox(0,0)[bl]{(c) \(t=\tau_{f}\) (forecast)}}


\put(05,23){\oval(10,6)}
\put(0.5,21.5){\makebox(0,0)[bl]{\(X_{2}(3)\)}}

\put(25,23){\oval(10,6)}
\put(21,21.5){\makebox(0,0)[bl]{\(Z_{2}(3)\)}}

\put(15,33){\oval(10,6)}
\put(11,31.5){\makebox(0,0)[bl]{\(Y_{2}(3)\)}}

\put(15,13){\oval(10,6)}
\put(10.5,11.5){\makebox(0,0)[bl]{\(H_{2}(1)\)}}

\put(11,23.5){\makebox(0,0)[bl]{\(3\)}}

\put(15.6,25){\makebox(0,0)[bl]{\(3\)}}

\put(09,27){\makebox(0,0)[bl]{\(3\)}}

\put(19,27){\makebox(0,0)[bl]{\(3\)}}

\put(09,17){\makebox(0,0)[bl]{\(2\)}}

\put(19,17){\makebox(0,0)[bl]{\(2\)}}


\put(10,23){\line(1,0){10}} \put(15,16){\line(0,1){14}}

\put(09.5,25){\line(1,1){05}} \put(20.5,25){\line(-1,1){05}}

\put(09.5,21){\line(1,-1){05}} \put(20.5,21){\line(-1,-1){05}}

\end{picture}
\end{center}

 Note, for the basic time point (\(\tau_{0}\)),
 two subsystems (as composite bottlenecks)
 have obtained  (Fig. 9):
 ~\(B_{2} =X_{2} \star Z_{2} \star H_{2}\),
 \(N(B_{2}) = (2;0,0,3)\);
 ~\(B_{4} = Y_{2} \star Z_{2} \star H_{2}\),
  \(N(B_{4}) = (3;0,2,1)\).
 For next time points,
 the following poset-like estimates are obtained:
 (i) \(t=\tau_{1}\):~
 \(N(B^{\tau_{1}}_{1}) = (3;0,1,2)\),
 \(N(B^{\tau_{1}}_{2}) = (2;0,1,2)\),
 \(N(B^{\tau_{1}}_{3}) = (2;0,2,1)\),
 \(N(B^{\tau_{1}}_{4}) = (2;0,2,1)\);
 (ii) \(t=\tau_{f} (\tau_{2})\):~
 \(N(B^{\tau_{2}}_{1}) = (3;0,0,3)\),
 \(N(B^{\tau_{2}}_{2}) = (3;1,1,1)\),
 \(N(B^{\tau_{2}}_{3}) = (2;1,0,2)\),
 \(N(B^{\tau_{2}}_{4}) = (2;1,1,1)\).
 As a result, the following subsystems
 are obtained as composite bottlenecks:

 (a) \(t=\tau_{1}\) (Fig. 14a):~
 ~\(B^{\tau_{1}}_{1} =X_{2} \star Y_{2} \star Z_{2}\),
 \(N(B^{\tau_{1}}_{1}) = (3;0,1,2)\);

 (b) \(t=\tau_{f} (\tau_{2})\) (Fig. 14b):~
 ~\(B^{\tau_{2}}_{1} =X_{2} \star Y_{2} \star Z_{2}\),
 \(N(B^{\tau_{2}}_{1}) = (3;0,0,3)\).

 Thus, the forecast bottleneck is: ~\(B^{\tau_{2}}_{1} = X_{2} \star Y_{2} \star Z_{2}\).
 Fig. 15 depicts
 a trajectory of the bottleneck.



\begin{center}
\begin{picture}(63,59)

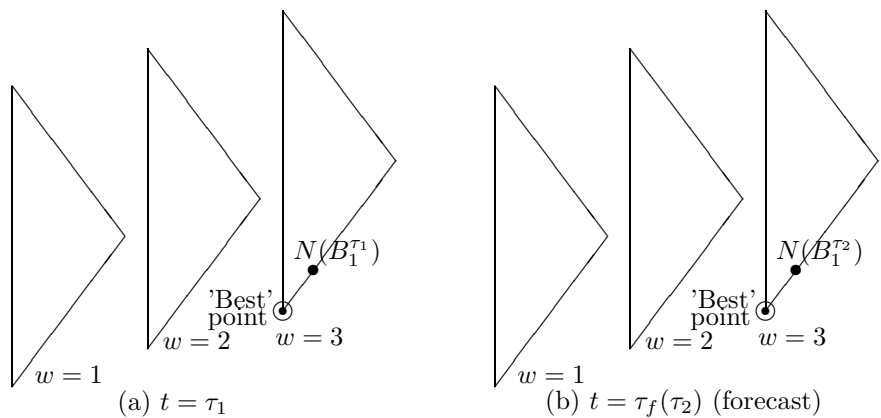
\put(020,00){\makebox(0,0)[bl]{Fig. 14. Poset-like scale for
 subsystem \(B\)}}

\put(14,05){\makebox(0,0)[bl]{(a) \(t=\tau_{1}\)}}

\put(00,09){\line(0,1){40}} \put(00,09){\line(3,4){15}}
\put(00,49){\line(3,-4){15}}

\put(18,14){\line(0,1){40}} \put(18,14){\line(3,4){15}}
\put(18,54){\line(3,-4){15}}

\put(36,19){\line(0,1){40}} \put(36,19){\line(3,4){15}}
\put(36,59){\line(3,-4){15}}


\put(40,24.5){\circle*{1.5}}
\put(37.4,25){\makebox(0,0)[bl]{\(N(B^{\tau_{1}}_{1})\)}}



\put(36,19){\circle*{1}} \put(36,19){\circle{2.5}}

\put(26,19.5){\makebox(0,0)[bl]{'Best'}}
\put(26,16.5){\makebox(0,0)[bl]{point}}

\put(03,09.5){\makebox(0,0)[bl]{\(w=1\)}}
\put(20,14){\makebox(0,0)[bl]{\(w=2\)}}
\put(35,14.5){\makebox(0,0)[bl]{\(w=3\)}}

\end{picture}
%
\begin{picture}(41,59)

\put(06.7,05){\makebox(0,0)[bl]{(b) \(t=\tau_{f} (\tau_{2})\)
(forecast)}}

\put(00,09){\line(0,1){40}} \put(00,09){\line(3,4){15}}
\put(00,49){\line(3,-4){15}}

\put(18,14){\line(0,1){40}} \put(18,14){\line(3,4){15}}
\put(18,54){\line(3,-4){15}}

\put(36,19){\line(0,1){40}} \put(36,19){\line(3,4){15}}
\put(36,59){\line(3,-4){15}}


\put(40,24.5){\circle*{1.5}}
\put(37.4,25){\makebox(0,0)[bl]{\(N(B^{\tau_{2}}_{1})\)}}


\put(36,19){\circle*{1}} \put(36,19){\circle{2.5}}

\put(26,19.5){\makebox(0,0)[bl]{'Best'}}
\put(26,16.5){\makebox(0,0)[bl]{point}}

\put(03,09.5){\makebox(0,0)[bl]{\(w=1\)}}
\put(20,14){\makebox(0,0)[bl]{\(w=2\)}}
\put(35,14.5){\makebox(0,0)[bl]{\(w=3\)}}

\end{picture}
\end{center}

\begin{center}
\begin{picture}(71,22)
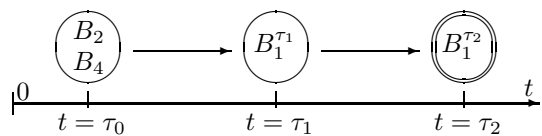
\put(00,00){\makebox(0,0)[bl] {Fig. 15. Trajectory of composite
bottlenecks}}

\put(0.5,9.5){\makebox(0,8)[bl]{\(0\)}}
\put(00,7.5){\line(0,1){3}}

\put(00,9){\vector(1,0){70}} \put(68,10){\makebox(0,8)[bl]{\(t\)}}

\put(6,5){\makebox(0,8)[bl]{\(t=\tau_{0}\)}}
\put(31,5){\makebox(0,8)[bl]{\(t=\tau_{1}\)}}
\put(56,5){\makebox(0,8)[bl]{\(t=\tau_{2}\)}}

\put(10,8.5){\line(0,1){2}} \put(35,8.5){\line(0,1){2}}
\put(60,8.5){\line(0,1){2}}

\put(8,17){\makebox(0,8)[bl]{\(B_{2}\)}}
\put(8,13){\makebox(0,8)[bl]{\(B_{4}\)}}

\put(10,16.5){\oval(08.5,9.5)} \put(16,16){\vector(1,0){13}}


\put(32,15){\makebox(0,8)[bl]{\(B^{\tau_{1}}_{1}\)}}

\put(35,16.5){\oval(08.5,9.5)} \put(41,16){\vector(1,0){13}}


\put(57,15){\makebox(0,8)[bl]{\(B^{\tau_{2}}_{1}\)}}

\put(60,16.5){\oval(08.5,9.5)} \put(60,16.5){\oval(07.5,8.5)}

\end{picture}
\end{center}

 Generally,
 it is reasonable to examine for the detection of the composite system bottlenecks
 an approach  'clique-based fusion based on
 graph streams'
 that was presented  in
  \cite{lev12clique}.

\section{Conclusion}

 The paper describes basic approaches to detection of bottlenecks
 in composite (modular) systems.
 In general, detection of system bottlenecks has to be used as a
 first stage of the system improvement/development process.
 The described approaches to detection of system bottlenecks
 are significant preliminary stage for the analysis and new design
 (redesign) of various systems.
 On the other hand, the considered approaches to detection of
 system bottlenecks are very close to
 system testing procedures
 (e.g., multi-function system testing \cite{lev12clique,levlast06}).
 In the future, it may be  reasonable to consider
 the following research directions:
 (1) examination of various  real-world applications;
 (2) examination of multi-stage frameworks for detection of system
 bottlenecks;
 (3) examination of system bottlenecks as system component(s)
 'trajectories';
 (4)  additional study of detection of bottlenecks in
  hierarchical (multi-layer) networks, for example:~
  (i) detection of low quality layer topology,
  (ii) detection of low quality
  connection between nodes of neighbor layers,
  (iii) detection of wrong or low quality assignment of nodes into
  structure/network layers.
%
%
 (5) taking into account uncertainty;
 and
 (6) usage of the described system approaches in
 education (computer science, engineering, applied mathematics).

\end{document}